\documentclass{article}

\PassOptionsToPackage{numbers, compress}{natbib}
\usepackage[final]{neurips_2025}
\usepackage[utf8]{inputenc} % allow utf-8 input
\usepackage[T1]{fontenc}    % use 8-bit T1 fonts
\usepackage[hidelinks]{hyperref}       % hyperlinks
\usepackage{url}            % simple URL typesetting
\usepackage{booktabs}       % professional-quality tables
\usepackage{amsfonts}       % blackboard math symbols
\usepackage{nicefrac}       % compact symbols for 1/2, etc.
\usepackage{microtype}      % microtypography
\usepackage{xcolor}         % colors
\usepackage{amsmath,amsfonts,bm}

\def\eqref#1{equation~\ref{#1}}
\def\1{\bm{1}}

\DeclareMathAlphabet{\mathsfit}{\encodingdefault}{\sfdefault}{m}{sl}
\SetMathAlphabet{\mathsfit}{bold}{\encodingdefault}{\sfdefault}{bx}{n}

\usepackage{comment}
\usepackage{xspace}
\usepackage{xcolor}
\usepackage{booktabs}

\usepackage{graphicx}
\usepackage{balance}  % for  \balance command ON LAST PAGE  (only there!)

\usepackage{graphics}
\usepackage{xspace}

\usepackage{booktabs}
\usepackage{soul}
\usepackage{balance}

\usepackage{float}
\usepackage{subcaption}

\usepackage{algorithm}
\usepackage[noend]{algpseudocode}
\usepackage{amsmath}
\usepackage{amssymb}
\usepackage{mathtools}
\usepackage{amsfonts}
\usepackage{amsbsy}
\usepackage{amsthm}
\usepackage[]{caption}
\usepackage{multirow}
\usepackage[mathscr]{eucal} %% For \mathscr
\usepackage{mathtools}
\usepackage{verbatim}
\usepackage{tabularx}
\usepackage{enumitem}
\usepackage{array}
\usepackage{soul}
\usepackage{dsfont}
\usepackage{url}
\usepackage{bbm}
\usepackage{listings}
\lstdefinestyle{textblock}{
    basicstyle=\ttfamily\small,
    backgroundcolor=\color{gray!5},
    frame=single,
    breaklines=true,
    showstringspaces=false,
    columns=fullflexible
}

\usepackage[most]{tcolorbox}

\usepackage{pifont}
\usepackage{xcolor,colortbl}
\usepackage[inkscapeformat=png]{svg}
\usepackage{makecell}
\usepackage{xspace}
\usepackage{footmisc} %[flushmargin]
\usepackage{wrapfig}
\hypersetup{
    colorlinks=true,
    linkcolor=black,
    citecolor = black,
    filecolor=magenta,      
    urlcolor=magenta,
    pdftitle={RDB2G-Bench},
    pdfpagemode=FullScreen,
    }

\definecolor{googleblue}{RGB}{66,133,244}
\definecolor{googlered}{RGB}{219,68,55}
\definecolor{googleyellow}{RGB}{244,180,0}
\definecolor{googlegreen}{RGB}{15,157,88}

\newtcolorbox{mybox}[2][]{%
  attach boxed title to top left
               = {yshift=-5pt},
  colback      = cyan!5!white,
  colframe     = googleblue,
  fonttitle    = \bfseries,
  colbacktitle = googleblue,%cyan!85!black,
  title        = #2,#1,
  enhanced,
}

\newtheorem{definition}{\textbf{Definition}}

\makeatletter
\newtheorem*{rep@theorem}{\rep@title}
\newcommand{\newreptheorem}[2]{%
\newenvironment{rep#1}[1]{%
 \def\rep@title{#2 \ref{##1}}%
 \begin{rep@theorem}}%
 {\end{rep@theorem}}}
\makeatother

\newreptheorem{theorem}{Theorem}

\definecolor{sunwoogreen}{rgb}{0.66, 0.89, 0.63}
\definecolor{sunwoogreen2}{RGB}{67, 148, 58}
\definecolor{sunwooyellow}{rgb}{1.0, 1.0, 0.0}
\definecolor{sunwooyellow2}{RGB}{228, 208, 10}

\newcommand\red[1]{\textcolor{red}{#1}}

\newcommand{\relbench}{\textsc{RelBench}\xspace}
\newcommand{\xhdr}[1]{\paragraph{#1.}\xspace}
\newcommand{\fone}{\texttt{rel-f1}\xspace}
\newcommand{\driverPosition}{\texttt{driver-position}\xspace}

\newcommand{\driverDNF}{\texttt{driver-dnf}\xspace}
\newcommand{\driverTopThree}{\texttt{driver-top3}\xspace}

\newcommand{\event}{\texttt{rel-event}\xspace}
\newcommand{\userAttendance}{\texttt{user-attendance}\xspace}
\newcommand{\userIgnore}{\texttt{user-ignore}\xspace}
\newcommand{\userRepeat}{\texttt{user-repeat}\xspace}

\newcommand{\avito}{\texttt{rel-avito}\xspace}
\newcommand{\userClick}{\texttt{user-clicks}\xspace}
\newcommand{\userVisit}{\texttt{user-visits}\xspace}
\newcommand{\userAdVisit}{\texttt{user-ad-visit}\xspace}
\newcommand{\adsCTR}{\texttt{ad-ctr}\xspace}

\newcommand{\stackex}{\texttt{rel-stack}\xspace}
\newcommand{\postPostLinked}{\texttt{post-post-related}\xspace}

\newcommand{\trials}{\texttt{rel-trial}\xspace}
\newcommand{\studyOutcome}{\texttt{study-outcome}\xspace}

\newcommand{\mr}[2]{\multirow{#1}{*}{#2}}
\newcommand{\mc}[3]{\multicolumn{#1}{#2}{#3}}

\newcommand{\ourdata}{\texttt{RDB2G-Bench}\xspace}
\newcommand{\baseline}{AR2N}
\newcommand{\github}{\url{https://github.com/chlehdwon/RDB2G-Bench}}
\newcommand{\huggingface}{\url{https://huggingface.co/datasets/kaistdata/RDB2G-Bench}}

\definecolor{case_gray}{RGB}{174, 200, 193}
\definecolor{case_lightblue}{RGB}{172, 233, 245}
\definecolor{case_green}{RGB}{16, 200, 113}
\definecolor{case_yellow}{RGB}{252, 207, 65}
\definecolor{case_pink}{RGB}{245, 103, 118}
\definecolor{case_orange}{RGB}{253, 149, 119}
\definecolor{case_blue}{RGB}{124, 200, 250}
\definecolor{case_lightgreen}{RGB}{165, 239, 85}
\definecolor{case_purple}{RGB}{200, 167, 249}

\definecolor{googleblue}{HTML}{4285F4}
\definecolor{googlered}{HTML}{EA4335}
\definecolor{googlegreen}{HTML}{34A853}
\definecolor{googleyellow}{HTML}{FBBC04}
\definecolor{googlepurple}{HTML}{8E7CC3}

\AtBeginDocument{%
  \providecommand\BibTeX{{%
    \normalfont B\kern-0.5em{\scshape i\kern-0.25em b}\kern-0.8em\TeX}}}

\title{RDB2G-Bench: A Comprehensive Benchmark for Automatic Graph Modeling of Relational Databases}

\author{
  Dongwon Choi, Sunwoo Kim, Juyeon Kim, Kyungho Kim, \\
  \textbf{Geon Lee, Shinhwan Kang, Kijung Shin}\\
   Kim Jaechul Graduate School of AI, KAIST\\
  \texttt{\{cookie000215, kswoo97, juyeonkim, kkyungho,} \\
  \texttt{geonlee0325, shinhwan.kang, kijungs\}@kaist.ac.kr} \\
   \And
  Myunghwan Kim \\
  Kumo.AI \\
  \texttt{myunghwan@kumo.ai} \\
}

\begin{document}

\maketitle
\begin{abstract}
  Recent advances have demonstrated the effectiveness of graph-based machine learning on relational databases (RDBs) for predictive tasks. 
Such approaches require transforming RDBs into graphs, a process we refer to as \textbf{RDB-to-graph modeling}, where rows of tables are represented as nodes and foreign-key relationships as edges.
Yet, effective modeling of RDBs into graphs remains challenging.
Specifically, there exist numerous ways to model RDBs into graphs, and performance on predictive tasks varies significantly depending on the chosen graph model of RDBs.
In our analysis, we find that the best-performing graph model can yield up to a 10\% higher performance compared to the common heuristic rule for graph modeling, which remains non-trivial to identify.
To foster research on intelligent RDB-to-graph modeling, we introduce \texttt{RDB2G-Bench}\xspace, the first benchmark framework for evaluating such methods.
We construct extensive datasets covering \textbf{5 real-world RDBs and 12 predictive tasks, resulting in around 50k graph model–performance pairs} for efficient and reproducible evaluations.
Thanks to our precomputed datasets, we were able to \textbf{benchmark 10 automatic RDB-to-graph modeling methods on the $12$ tasks about 380$\times$ faster} than on-the-fly evaluation, which requires repeated GNN training.
Our analysis of the datasets and benchmark results reveals key structural patterns affecting graph model effectiveness, along with practical implications for effective graph modeling.
Our datasets and code are available at \url{https://github.com/chlehdwon/RDB2G-Bench}.
\end{abstract}

\section{Introduction}
\label{sec:introduction}
A relational database (RDB) is a collection of data organized into multiple tables connected by shared keys.
RDBs enable systematic and efficient management of related information through query languages such as SQL, and have been widely adopted across diverse industries, including finance~\citep{ionescu2024assessment}, healthcare~\citep{tarigan2023relational}, and e-commerce~\citep{saweczko2023comparative}.
This widespread use has led to the emergence of diverse machine learning applications built upon RDBs.

For machine learning on RDBs, recent studies have explored graph-based approaches~\cite{fey2023relational, gan2024graph, robinson2024relbench, wang20244dbinfer, yuan2025contextgnn, chen2025relgnn}, which involve modeling RDBs as graphs.
Typically, rows of tables are modeled as nodes in the resulting graph, and foreign key (FK) relationships are represented as edges.
Graph modeling, where graph neural networks are subsequently applied, effectively captures structural dependencies and leads to improved performance across a range of machine learning tasks~\cite{cvitkovic2020supervisedrd, robinson2024relbench, wang20244dbinfer}.

One important consideration in graph-based approaches is the variety of ways to model RDBs as graphs.
For example, a single table row representing a user transaction on an item can be modeled either as (1) a node representing the transaction or (2) an edge linking the corresponding user and item.
Additionally, one may choose to include only a subset of tables (i.e., nodes) and FK relationships (i.e., edges), excluding those considered less relevant.
We call the process of representing RDBs as graphs \textbf{RDB-to-graph modeling}, and refer to the resulting graph representation as a \textbf{graph model}.

\begin{figure*}[t]
    \vspace{-0.5mm}
    \centering
    {\includegraphics[width=0.95\linewidth]{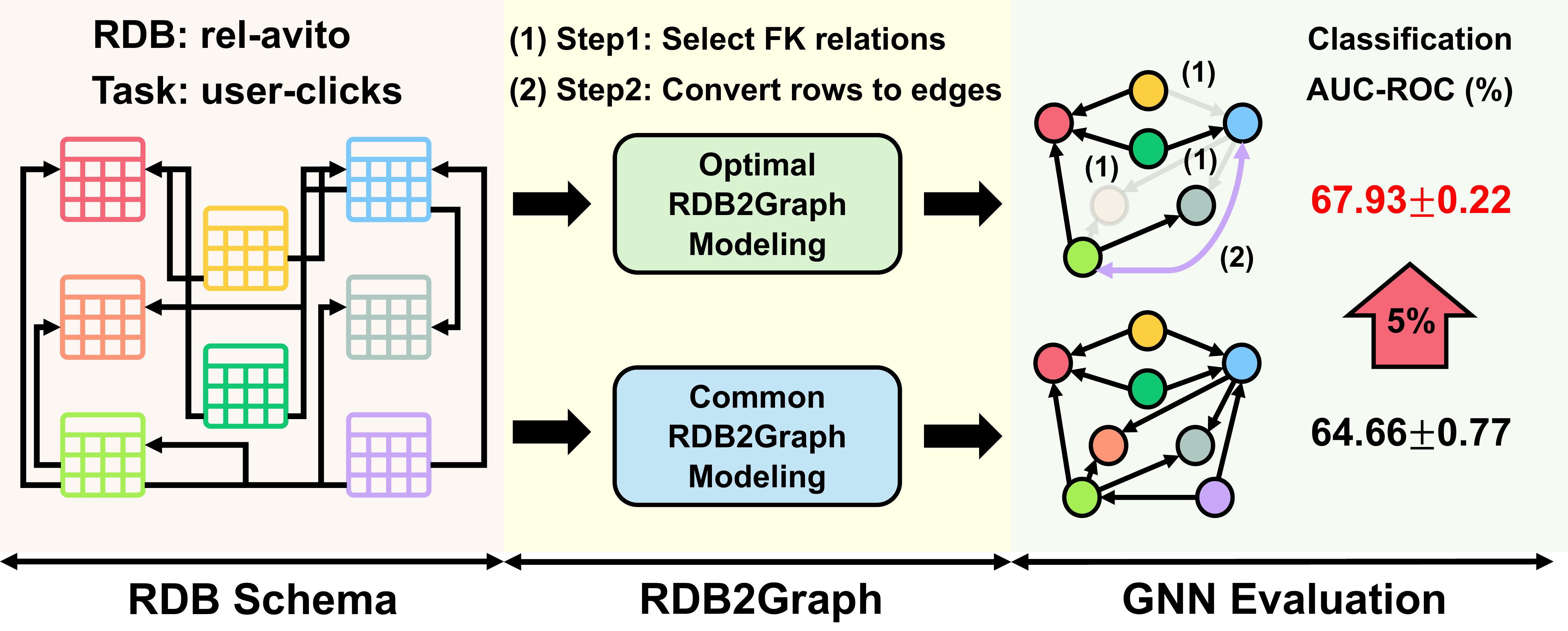}}
    \caption{
    Overview of key concepts. An RDB schema is converted into various network schemas using different RDB-to-graph (RDB2Graph) modeling methods.
    Graphs are then constructed from these schemas, where graph neural networks (GNNs) are trained and evaluated.
    In the given example task, optimal modeling yields up to a 5\% performance improvement over a widely-used heuristic~\cite{fey2023relational}.
    Note that the optimal graph model selectively uses tables and foreign key (FK) relations, with table rows modeled as edges, while the heuristic models the entire RDB with all table rows as nodes.
    }
    \label{fig:introexample}
\end{figure*}

Careful RDB-to-graph modeling is crucial since the empirical effectiveness of graph-based approaches heavily depends on the choice of graph model.
For example, our analysis of a graph-based approach~\cite{fey2023relational, robinson2024relbench} on the \avito dataset shows that modeling the rows of certain tables as edges (rather than nodes) and excluding some FK relations led to a downstream-task performance gain of up to 5\% compared to using a commonly-used fixed modeling rule~\cite{fey2023relational}, as illustrated in Figure~\ref{fig:introexample}.

Despite the importance of intelligent RDB-to-graph modeling, research in this area remains in its early stages, with one main reason being the difficulty of evaluation.
First, there are numerous possible graph models (see Figure~\ref{fig:datasetsummary}(a) for examples from real-world RDBs),
making exhaustive comparisons extremely expensive.
Moreover, evaluating even a single graph model is computationally costly, as it typically involves training graph neural networks (GNNs) on a large graph.

To facilitate research on intelligent RDB-to-graph modeling, we introduce \ourdata, the first RDB-to-graph modeling benchmark framework, with the following key contributions:

\begin{itemize}[leftmargin=*]
    \item \textbf{Datasets.} 
    \ourdata provides precomputed performance metrics (training, validation, and test performance and runtime over at least five trials)  for \textbf{50k graph models}
    based on $5$ real-world RDBs and 12 predictive tasks.
    They enable researchers to evaluate their modeling methods without training GNNs on the resulting graph models.
    \item \textbf{Benchmarks.} 
    We present the extensive benchmark results of 10 automatic RDB-to-graph modeling methods across 12 tasks. Thanks to our precomputed datasets, we were able to obtain the benchmark evaluation results about \textbf{380$\times$ faster} than on-the-fly evaluation with repeated GNN training.
    \item \textbf{Observations.} Our analysis of the datasets and benchmark results identifies important factors that impact performance, providing practical insights for effective graph modeling.

\end{itemize}

Our datasets and code are available at: \url{https://github.com/chlehdwon/RDB2G-Bench}.

\section{Related Works}
\label{sec:relatedwork}

\subsection{Relational Deep Learning}\label{subsec:rdl}
Early approaches to relational database (RDB) learning relied on feature engineering~\cite{kaggle-survey} (e.g., table joins), transforming an RDB into a single table to leverage tabular machine learning \cite{chen2016xgboost, ke2017lightgbm, arik2021tabnet, gorishniy2021revisiting}. 
Recent works on relational deep learning (i.e., deep learning on RDBs)~\cite{cvitkovic2020supervisedrd, schlichtkrull2018modeling, fey2023relational, gan2024graph, dwivedi2025relational} pioneered modeling RDBs as graphs,
which facilitate the capture of cross-table dependencies. Building on these graph models, advanced methods~\cite{chen2025relgnn,yuan2025contextgnn} have achieved improved predictive performance on RDBs.

\subsection{Benchmarks on Relational Deep Learning}\label{subsec:rdlbench}
Early benchmarks (e.g., the CTU relational learning repository~\cite{motl2015ctu}) set the stage for evaluating machine learning (ML) models on real-world RDBs.
Recent benchmarks (e.g., RDBench~\cite{zhang2023rdbench}, RelBench~\cite{robinson2024relbench}, and 4DBInfer~\cite{wang20244dbinfer}) shift the focus to graph-based approaches, which often show improved performance over traditional tabular learning methods~\cite{ke2017lightgbm, chen2016xgboost}.

% Different points of our work from previous works
\textbf{Distinctive property of our benchmark:} Existing benchmarks provide RDBs and predictive tasks to compare machine learning (ML) methods, which are typically based on the same graph presentation of the RDBs.
Our benchmark, \ourdata, however, is designed to evaluate RDB-to-graph modeling strategies, which provide graph models based on which various modeling methods perform.

Specifically, by offering precomputed performance metrics for 50k graph models, \ourdata enables comparing various graph modeling strategies without repeated GNN training (see Section~\ref{subsec:benchmarkresults} for efficiency gains).

\subsection{Graph-based Modeling of RDBs}\label{subsec:rdb2graphsearch}
% Prior methods heavily rely on heuristic methods to convert RDBs to appropriate forms.
Traditionally, ML on RDBs has relied heavily on heuristic methods for automating schema transformations.
These methods include rule-based relationship mining~\cite{abedjan2015profiling, yao2008mining, liu2010discover, koutras2020rema} and conversion of RDBs into single-table formats through predefined aggregation functions~\cite{kanter2015deep, zhu2017auto}. In addition,
recent studies have focused on graph modeling, such as mapping table rows to nodes (Row2Node)~\cite{robinson2024relbench} or to nodes or edges (Row2N/E)~\cite{wang20244dbinfer}. However, heuristic-driven approaches often produce suboptimal RDB representations (see Figure~\ref{fig:introexample} for an example).
More recently, AutoG~\cite{chen2025autog} was proposed, leveraging large language models (LLMs) to actively explore effective graph models of RDBs.

Our benchmark, \ourdata, enables the evaluation of various RDB-to-graph modeling methods, including both heuristic and LLM-based approaches (see Section~\ref{subsec:benchmarksettings} for the ten methods included in the benchmark), ultimately fostering the development of more advanced modeling strategies.

\section{Dataset Design for \ourdata}
\label{sec:datasetdesign}
In this section, we introduce the datasets provided in \ourdata, our benchmark for the
RDB-to-graph modeling problem. The problem is defined as follows:

\begin{mybox}[colback=googleblue!10!white,colframe=googleblue]{Problem Definition.}\label{box:claim}%[colback=cyan!5!white,colframe=teal!35!black]
  \begin{definition}[RDB-to-Graph Modeling] \label{def:prolem}
    ~
    \begin{itemize}[leftmargin=*]
        \item \textbf{Given:} A relational database $\mathcal{R}$, and a graph neural network $\mathcal{M}$ for a downstream task $\mathcal{T}$.
        \item To \textbf{Find:} the graph model $\mathcal{G}^*$ of $\mathcal{R}$
         that maximizes the performance $\mathcal{P}$ of $\mathcal{M}$ trained and evaluated on  $\mathcal{T}$, i.e.,
        \[
        \mathcal{G}^* = \arg\max\nolimits_{\mathcal{G}\in \mathcal{F}(\mathcal{R}, \mathcal{T})} \mathcal{P}(\mathcal{M}(\mathcal{G}), \mathcal{T}),
        \]
    where $\mathcal{F}(\mathcal{R}, \mathcal{T})$
    is the set of possible graph models of $\mathcal{R}$ on $\mathcal{T}$, defined in
    Section~\ref{subsec:searchspace}; and example metrics of $\mathcal{P}$ are provided in Section~\ref{subsec:datasettings}.
    \end{itemize}
    \end{definition}
\end{mybox}

Each dataset comprises graph models (i.e., graph-structured representations) of a given relational database (RDB)
paired with downstream task performance metrics of graph neural networks (GNNs) (denoted by $\mathcal{M}$) trained on those graph models.
We first describe the considered design space of graph models  (Section~\ref{subsec:searchspace}), followed by the procedure for obtaining GNN performance metrics on each graph model (Section~\ref{subsec:datasettings})

\textbf{RDBs and downstream predictive tasks.}
Our \ourdata datasets were created based on 5 real-world RDBs and 12 predictive tasks provided in RelBench~\cite{robinson2024relbench}.
The RDBs and tasks, summarized in Figure~\ref{fig:datasetsummary}(a), were carefully selected to cover a diverse set of tasks (classification, regression, and recommendation) and domains.
Refer to Appendix~\ref{app:relbenchdataset} for the details of the RDBs and tasks.

\subsection{Construction of Graph Models}\label{subsec:searchspace}

\begin{figure}[t]
  \centering
  \textbf{(a) Summary of the \ourdata datasets, which cover 50k graph models in total.}
  \vspace{2mm}
  \label{tab:searchspace}
  \setlength{\tabcolsep}{4pt}
  {
      \small
      \setlength{\tabcolsep}{3pt}
      \begin{tabular}{lllc|>{\columncolor{yellow!15}}c|ccc}
        \toprule
        \multirow{2}{*}{\textbf{RDB}} & \multirow{2}{*}{\textbf{Task Name}} & \multirow{2}{*}{\textbf{Type}} & \multirow{2}{*}{\textbf{\# Tables}} & \textbf{\# Graph} & \multicolumn{3}{c}{\textbf{Performance Statistics}} \\
        & & & & \textbf{Models} & \textbf{Best} & \textbf{AR2N~\cite{robinson2024relbench}} & \textbf{Worst} \\
        \midrule
        \midrule
        \multirow{4}{*}{\texttt{rel-avito}} 
        & user-clicks (UC) & classification & \multirow{4}{*}{8} & 944 & 67.93 & 64.66 & 60.89 \\
        & user-visits (UV) & classification &  & 944 & 66.33 & 65.97 & 59.83 \\
        & ad-ctr (AC) & regression &  & 1304 & 0.039 & 0.040 & 0.044 \\
        & user-ad-visit (UAV) & recommendation &  & 909 & 3.682 & 3.661 & 0.159 \\
        \midrule
        \multirow{3}{*}{\texttt{rel-event}} 
        & user-repeat (UR) & classification & \multirow{3}{*}{5} & 214 & 82.29 & 77.65 & 63.96 \\
        & user-ignore (UI) & classification &  & 214 & 82.82 & 82.22 & 74.29 \\
        & user-attendance (UA) & regression &  & 214 & 0.237 & 0.244 & 0.266 \\
        \midrule
        \multirow{3}{*}{\texttt{rel-f1}} 
        & driver-dnf (DD) & classification & \multirow{3}{*}{9} & 722 & 74.56 & 73.14 & 67.40 \\
        & driver-top3 (DT) & classification &  & 722 & 81.88 & 78.11 & 75.37 \\
        & driver-position (DP) & regression &  & 722 & 3.831 & 3.913 & 4.171 \\
        \midrule
        \texttt{rel-stack} & post-post-related (PPL) & recommendation & 7 & 7979 & 12.04 & 10.82 & 0.006 \\
        \midrule
        \texttt{rel-trial} & study-outcome (SO) & classification & 15 & 36863 & 70.91 & 68.09 & 62.85 \\
        \bottomrule
      \end{tabular}
  }
  \vspace{3mm}
  
  \textbf{(b) Visualization of the \ourdata datasets on three predictive tasks.}
  \vspace{3mm}
  \includegraphics[width=0.99\linewidth]{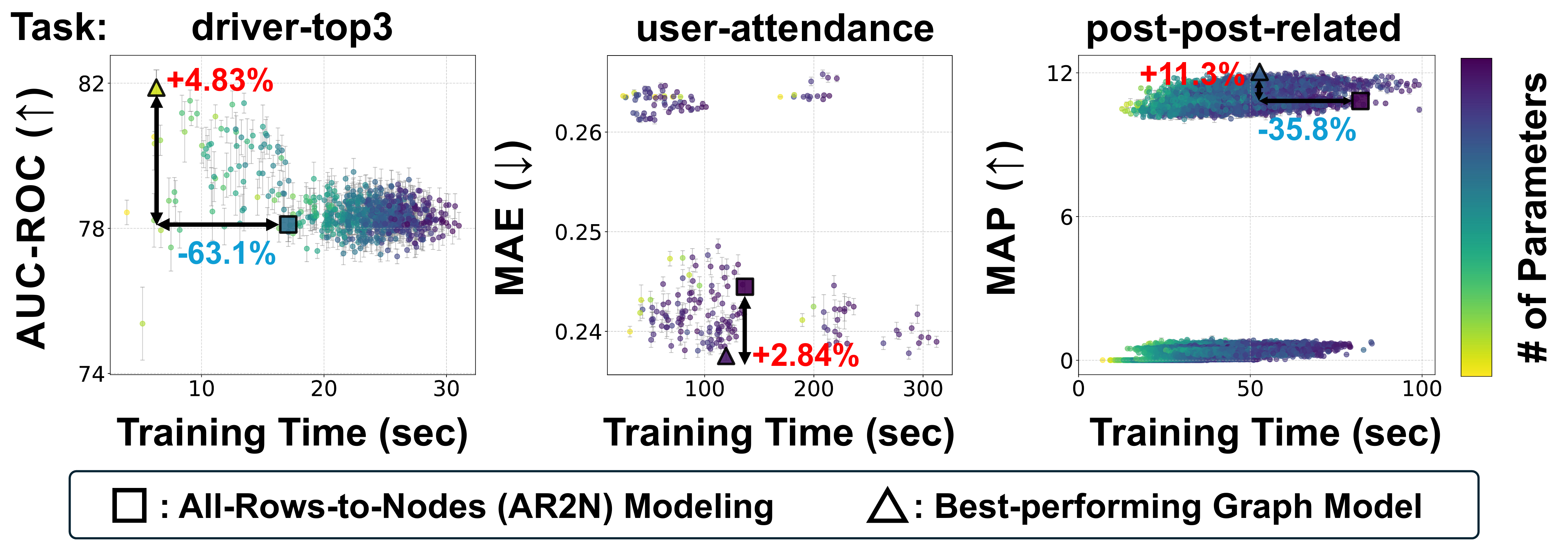}
  \caption{
        (a) 
        We summarize the RDBs, tasks, and their associated graph models.
    For each classification, regression, and recommendation task, we collect AUC-ROC (\%), MAE, and MAP (\%), respectively, on each graph model. For each task, we report the performances on the best graph model, the worst model, and that given by AR2N modeling~\cite{robinson2024relbench}.
    (b) For three tasks (\driverTopThree, \userAttendance, \postPostLinked),
    we visualize the distribution of performances on the downstream task (Y-axis) across all graph models, along with training time per epoch (X-axis) and the parameter size of the graph neural network (indicated by color).
    Note that there exist graph models yielding substantial improvements in both performance and efficiency compared to those generated by widely-used AR2N modeling~\cite{robinson2024relbench}.
    }
    \vspace{3mm}
    \label{fig:datasetsummary}
\end{figure}

As discussed in Section~\ref{sec:introduction}, there are various ways to model each RDB as a graph, and the performance of GNNs depends heavily on the chosen graph model.
In \ourdata, we consider the design choices in the following two steps of RDB-to-graph modeling, both of which have a significant impact on the downstream-task performance of GNNs, as shown in Section~\ref{sec:datasetanalysis}:
\begin{itemize}[leftmargin=*]
    \item \textbf{Step 1:} Selecting which tables and foreign key (FK) relationships to include in the graph model.
    \item \textbf{Step 2:} Selecting how to represent the rows of each table, as either nodes or edges, in the graph.
\end{itemize}
That is, for each RDB $\mathcal{R}$ (e.g., \texttt{rel-avito}), our dataset consists of graph models that result from different combinations
of design choices in \textbf{Step 1} and \textbf{Step 2}, which we denote by $\mathcal{F}(\mathcal{R}, \mathcal{T})$ in Definition~\ref{def:prolem} (refer to Figure~\ref{fig:introexample} for examples).
These graph models are used to train graph neural networks (GNNs) for predictive tasks (e.g., \texttt{user-clicks}) defined on the RDB (as described in Section~\ref{subsec:datasettings}), and to be used for this purpose, the graph models must satisfy several constraints.
First, a valid graph model must select the \textit{task table}, which the considered predictive task is defined on.
Second, all selected tables should be connected to the task table via a path whose length does not exceed the number of GNN layers. Violating this constraint would result in some nodes being unreachable during message passing, leading to degraded performance.
Third, (the rows of) a table can be modeled as edges only if the table has exactly two FKs and its primary key (PK) is not referenced by any FKs in other tables.
Note that, for a table with more than two FKs, hyperedge modeling would be required, which is beyond our scope, and edge modeling of a table whose PK is referenced by FK renders those FK relationships unrepresentable.
Also note that these constraints may lead to different graph model spaces for downstream tasks defined on the same RDB. As summarized in Figure~\ref{fig:datasetsummary}(a), we constructed about \textbf{50k graph models} spanning the aforementioned RDBs and downstream tasks; and evaluated their downstream task performance metrics, as described below.

\vspace{-1mm}
\subsection{Collection of Performance Metrics}\label{subsec:datasettings}
For each task and each constructed graph model, we collected the following performance metrics (i.e., $\mathcal{P}$ in Definition~\ref{def:prolem}):
\begin{itemize}[leftmargin=*]
    \item \textbf{Predictive performance:} The training, validation, and test performances on the downstream task.
    \item \textbf{Runtime:} The total elapsed time per epoch over at least five trials.
    \item \textbf{Parameter size:} The total number of learnable parameters of predictive GNNs, which depends on the graph model.
\end{itemize}
Specifically, we collected the performance metrics under the following setups:

\textbf{Machines.}
All experiments were conducted using NVIDIA RTX A6000 GPUs with 48GB of memory and Intel Xeon Silver 4210/4310 CPUs. Constructing all datasets required about 10,400 GPU hours.

\textbf{Predictive GNNs.}
As predictive machine learning models, we used GNNs provided by RelBench~\cite{robinson2024relbench}, which leverage PyTorch-Frame~\cite{hu2024pytorch} to encode each table as input to the GNNs.
Specifically, for classification and regression tasks, we employed Heterogeneous GraphSAGE~\cite{hamilton2017inductive} with sum aggregation to update embeddings for the final predictions. For recommendation tasks, we utilized ID-GNN~\cite{you2021identity}.
As the RelBench GNNs do not support edge modeling of tables, we extended them to incorporate edge features and used them when tables are modeled as edges (refer to Appendix~\ref{app:relbenchmodel}).
We additionally used three alternative predictive GNNs, as described in Section~\ref{subsec:modelgeneralization}.

\textbf{Training Details.}
Based on the training protocol provided by RelBench~\cite{robinson2024relbench} (including the training, validation, and test splits), we tuned only the learning rate for each graph model. 
The other hyperparameters were fixed to the combinations that yielded the best overall performance on 50 randomly sampled graph models. 
For classification and regression tasks, we selected the learning rate from \{0.005, 0.001, 0.0005\}, and for recommendation tasks, from \{0.001, 0.0005, 0.0001\}.
The number of training epochs was fixed at 20, which was empirically sufficient for convergence. For reliability, we repeated each experiment with 15 different random seeds for the \fone{} and \event{} datasets, and with 5 seeds for the remaining datasets.
Refer to Appendix~\ref{app:datasethyperparams} for further  details.

\section{Observations from the \ourdata Datasets} % Results
\label{sec:datasetanalysis}
In this section, we highlight key observations from our analysis of the constructed \ourdata datasets.
Unless otherwise stated, we use the predictive models described in Section~\ref{subsec:datasettings}.
\textbf{For the full analysis results omitted due to space constraints, refer to Appendix~\ref{sec:appendix_res}}.

\vspace{-3mm}
\subsection{Obs 1. ``Finding the best graph models is worthwhile, given their performance benefits.''
}\label{subsec:overallres}

{\textbf{Analysis Overview.}
We analyze the distribution of downstream-task performances across graph neural networks (GNNs) trained on different graph models of RDBs.
We also examine how performance relates to (1) mean GNN training time (per epoch) and (2) the number of GNN parameters.}

\textbf{Results.}
As shown in Figure~\ref{fig:datasetsummary}(b), the performance varies significantly across graph models. For example, in the \postPostLinked task, the performance difference spans 11.3\%.
Notably, for each task, there exist graph models that lead to significantly better performance (e.g., 4.83\% improvement in the 
\driverTopThree task)
than the one produced by the widely used all-rows-to-nodes (AR2N) modeling~\cite{robinson2024relbench}, and they often even require shorter training time and fewer parameters.

\textbf{Implications.}
These results suggest that it is not always beneficial to include all tables and foreign key (FK) relationships in a graph model.
Instead, selecting only the most relevant tables can lead to better performance, while also reducing GNN parameters and training time.
Moreover, there exist graph models that are both effective and efficient, yet difficult to identify using simple heuristics.

\subsection{Obs 2. ``Modeling table rows as edges can be crucial, depending on the task.''
}\label{subsec:row2necomparison}
\textbf{Analysis Overview.} 
We compare the two design choices for modeling the rows of a table in an RDB:
(1) \texttt{Row2Node}, which always represents table rows as nodes, and
(2) \texttt{Row2Edge}, which represents table rows as edges, especially when the table satisfies the constraints discussed in Section~\ref{subsec:searchspace}.
Specifically, we compare the downstream task performance distributions of the top 10\% highest-performing graph models, grouped by design choice.

\begin{figure*}[t]
    \vspace{-3mm}
    \centering
{\includegraphics[width=0.95\linewidth]{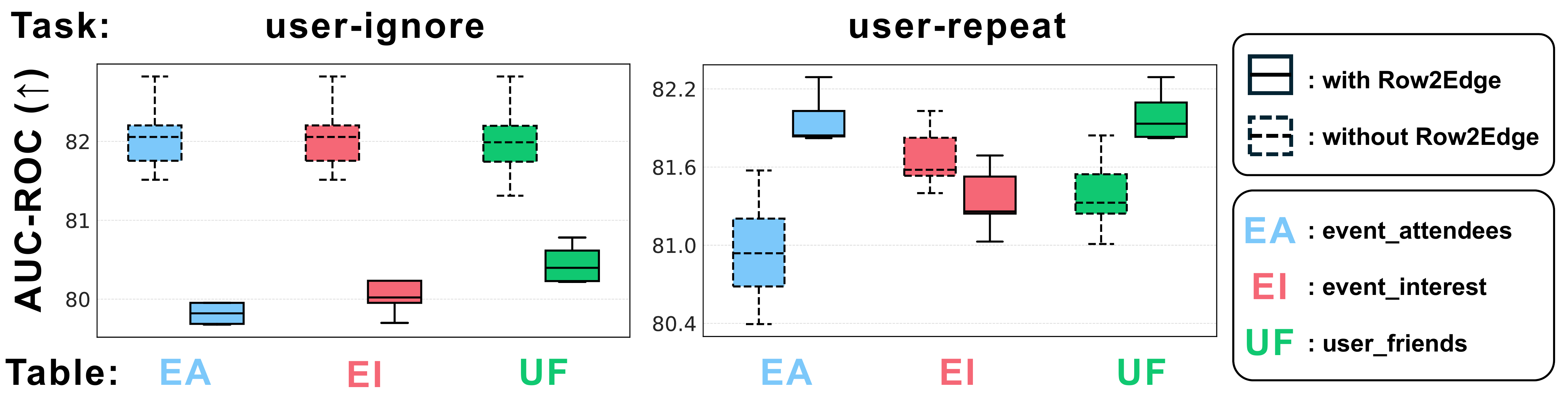}}
    \vspace{0.2cm}
    \caption{
        \textbf{Modeling table rows as edges (\texttt{Row2Edge}) can be crucial, depending on the task (\textbf{Obs 2}).} 
        \texttt{\textcolor{case_blue}{EA}} (event\_attendees), \texttt{\textcolor{case_pink}{EI}} (event\_interest), and \texttt{\textcolor{case_green}{UF}} (user\_friends) indicate the tables whose rows can be modeled as edges.
        Note that \texttt{Row2Edge} modeling improves performance for the \userRepeat task, but not for the \userIgnore task, even when both are defined on the same RDB. 
    }
    \label{fig:row2nemain}
\end{figure*}

\textbf{Results.}
Figure~\ref{fig:row2nemain} highlights two representative patterns.
For the \userIgnore task on the \event dataset, \texttt{Row2Edge} modeling consistently leads to performance degradation across all tables.
In contrast, for the \userRepeat task on the same RDB, 
the \texttt{Row2Edge} modeling of the \texttt{\textcolor{case_blue}{EA}} (event\_attendees) and \texttt{\textcolor{case_green}{UF}} (user\_friends) tables  significantly improves AUC-ROC scores.

\textbf{Implications.} 
These findings suggest that even within the same RDB, the benefit of \texttt{Row2Edge} modeling varies significantly across downstream tasks.
This variability suggests that no universal rule-of-thumb exists for modeling table rows; the choice should be carefully designed to achieve good performance on the target downstream tasks.
% Instead, selecting the best row modeling strategy must be done by task-specific considerations.

\subsection{Obs 3. ``Top-performing graph models share common substructures.''}\label{subsec:commonsubstructure}
\textbf{Analysis Overview.}
We investigate whether top-performing graph models share common substructures by examining the top 1\% best-performing graph models for each task.

\begin{figure*}[t]
    \vspace{-2mm}
    {\includegraphics[width=0.9\linewidth]{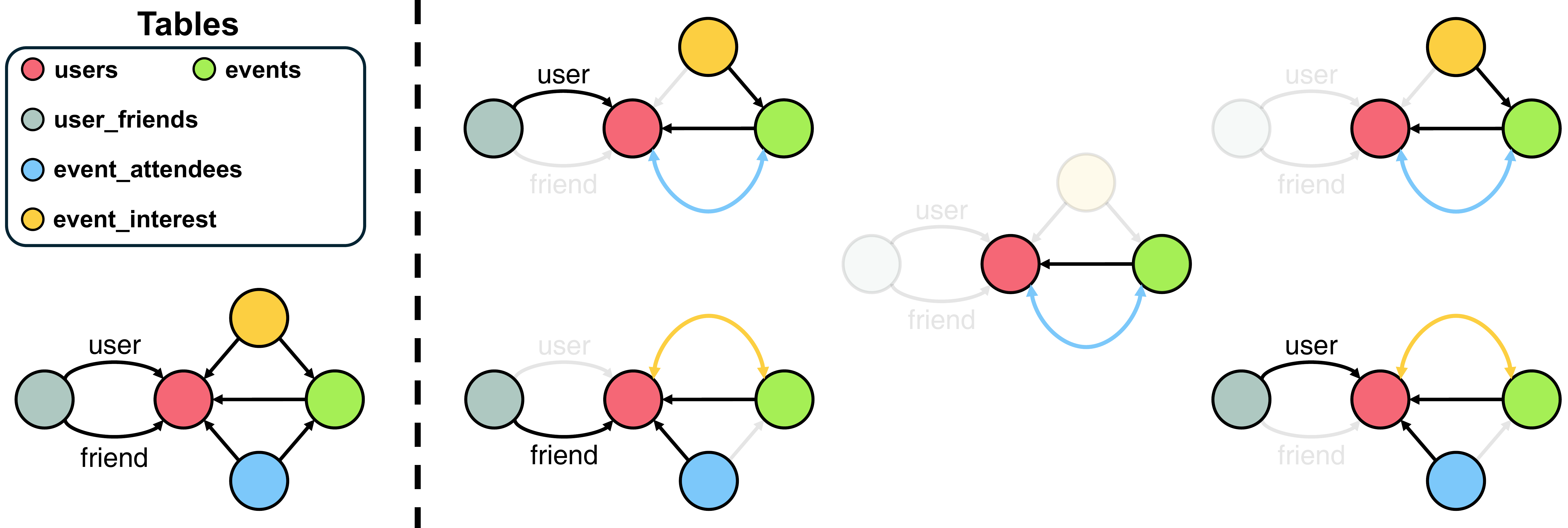}}
    \caption{\textbf{Top-performing graph models share common substructures (\textbf{Obs 3}).}
    As shown in their graph models, the top-5 graph models commonly (a) include the foreign-key (FK) relationship \textcolor{case_lightgreen}{events} $\rightarrow$ \textcolor{case_pink}{users} and (b) model either the \textcolor{case_blue}{event\_attendees} table or the \textcolor{case_yellow}{event\_interest} table as edges.
    Note that the \textcolor{case_gray}{users\_friends} table has two FKs (\texttt{user} and \texttt{friend}) both referencing the \textcolor{case_pink}{users} table.
    }
    \label{fig:caseeventua}
\end{figure*}

\textbf{Results.}
Figure~\ref{fig:caseeventua} shows a representative case from the \userAttendance task on the \event dataset. Note that all top-5 best performing graph models include the foreign-key relationship \textcolor{case_lightgreen}{events} $\rightarrow$ \textcolor{case_pink}{users}. They also model either the \textcolor{case_blue}{event\_attendees} table or the \textcolor{case_yellow}{event\_interest} table as edges.

\textbf{Implications.}
This empirical observation indicates the presence of \textbf{common substructures} that are critical for downstream task performance. Finding these structures can be key to identifying effective and efficient graph models.

\subsection{Obs 4. ``Different tasks may require different graph models, even on the same RDB.''}\label{subsec:taskvariation}
\textbf{Analysis Overview.} 
We analyze cross-task performance correlations to examine whether the effectiveness of a graph model in one downstream task implies its effectiveness in others.
Specifically, we measure the Spearman rank correlation~\cite{spearman1961proof} between performances on tasks within the same RDB.

\begin{figure*}[t]
    \vspace{-3mm}
    \centering
    {\includegraphics[width=0.95\linewidth]{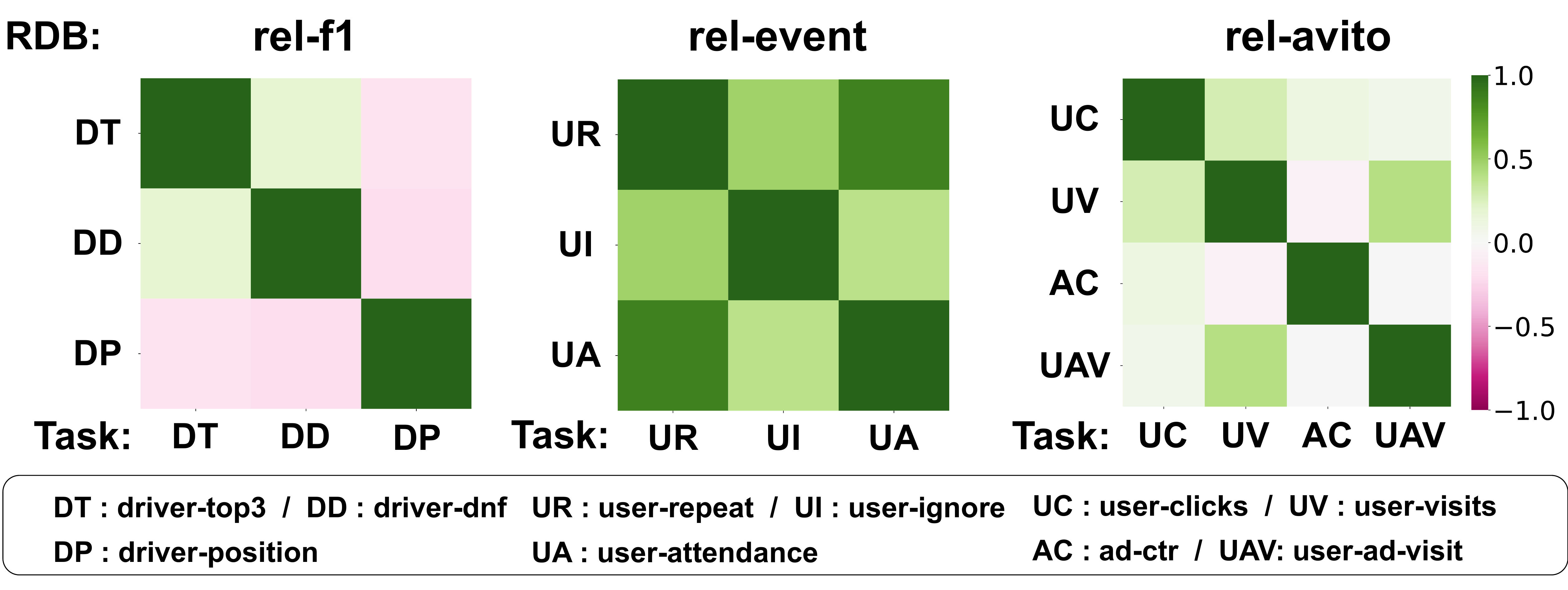}}
    \caption{
        \textbf{Different tasks may require different graph models, even on the same RDB (\textbf{Obs 4}).}
        Spearman correlations between downstream task performances on each RDB (\fone, \event, or \avito) are generally low (below 0.4), except for tasks with closely aligned goals.
    }
    \label{fig:analysistask}
\end{figure*}

\textbf{Results.}
As shown in Figure~\ref{fig:analysistask}, tasks whose objectives are highly aligned (spec., \userAttendance and \userRepeat in \event)\footnote{\userAttendance asks how many events each user will attend, and \userRepeat asks whether a user will repeat attendance at an event.} exhibit a high Spearman rank correlation exceeding 0.9. 
However, most task pairs exhibit low correlations (below 0.4), indicating that a graph model effective for one task may not generalize well to others, even within the same RDB.

\textbf{Implications.}
These results suggest that an effective RDB-to-graph modeling strategy should account not only for the characteristics of the RDB but also for those of the specific downstream task. Moreover, reusing a graph model is effective only across tasks with closely aligned objectives.

\subsection{Obs 5. ``Effectiveness of graph models generalizes across predictive GNNs.''
}\label{subsec:modelgeneralization}
\textbf{Analysis Overview.}
We analyze performance correlations across different predictive GNNs applied to the same graph models to examine whether a graph model effective with one GNN (e.g., Heterogeneous GraphSAGE~\cite{hamilton2017inductive}) is also effective with others.
To this end, we consider three additional GNNs (GraphSAGE with mean aggregation~\cite{hamilton2017inductive} and GIN~\cite{xu2019powerful}) and a graph transformer (GPS~\cite{rampavsek2022recipe})
For their details, refer to Appendix~\ref{app:relbenchmodel}.
As in the previous analysis, we employ the Spearman rank correlation \cite{spearman1961proof}.

\begin{figure*}[t]
    \centering
    {\includegraphics[width=0.95\linewidth]{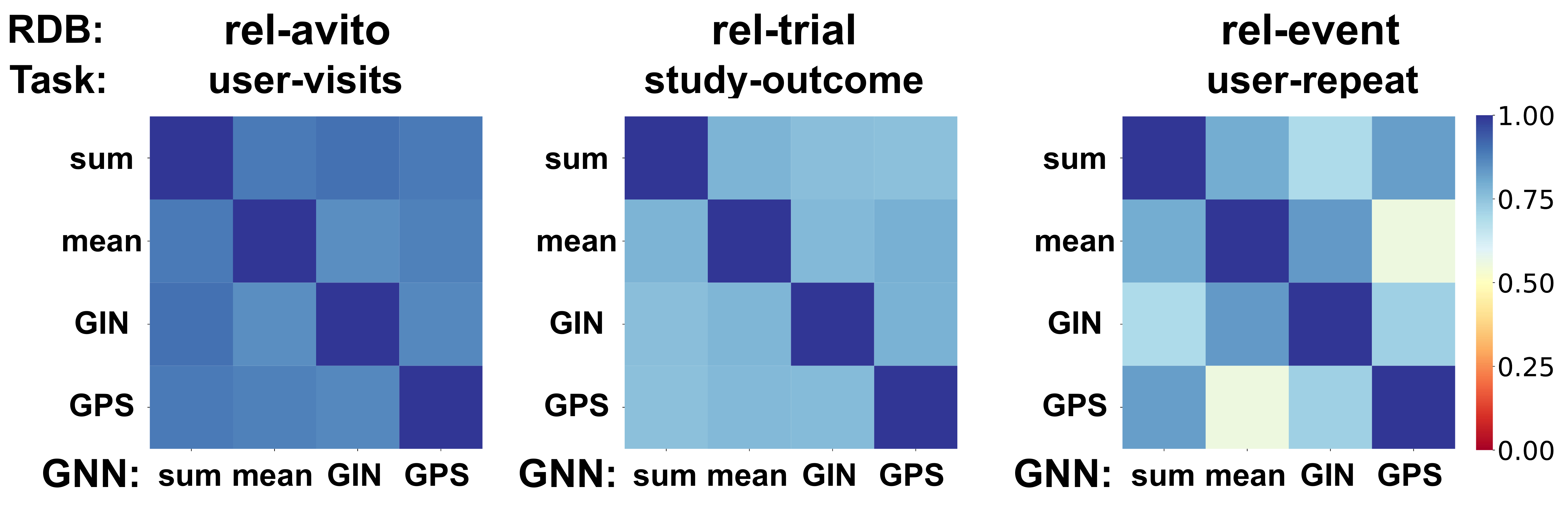}}
    \caption{
        \textbf{Effectiveness of graph models  generalizes across predictive GNNs (\textbf{Obs 5}).} Spearman correlations between different predictive GNNs—GraphSAGE with sum and mean aggregation (denoted as sum and mean), GIN, and GPS—are generally high (above 0.8) across the three RDBs (\avito, \trials, and \event).
    }
    \label{fig:modelgeneralization}
\end{figure*}

\textbf{Results.}
As shown in Figure~\ref{fig:modelgeneralization}, the correlations are generally high (above 0.7) in most cases, indicating that the effectiveness of graph models tends to generalize across different GNNs.
Notably, the graph transformer also exhibits strong correlations with GNNs, despite their architectural differences.

\textbf{Implications.}
This cross–GNN generalizability of graph models suggests that RDB-to-graph modeling strategies trained and shown to work well using our benchmark, \ourdata, can be effective across various predictive GNNs, highlighting the broad utility of \ourdata.

\section{Benchmark Results on the \ourdata Datasets}
\label{sec:benchmark}
In this section, we review the benchmark results of ten RDB-to-graph modeling methods on our \ourdata datasets. 

\subsection{Baselines for RDB-to-Graph Modeling}\label{subsec:benchmarksettings}
We benchmark ten RDB-to-graph modeling methods, which are categorized into (a) heuristic-based methods, (b) action-driven search algorithms, and (c) an LLM-based approach.
All methods are designed to select a graph model among those in our \ourdata datasets (see Section~\ref{subsec:searchspace} for details on our graph model space).

\textbf{Heuristic-Based Methods.}
The following heuristic-based methods rely on simple rules to model an RDB as a graph, without explicitly searching for effective graph models:
\begin{itemize}[leftmargin=*]
    \item \textbf{S1. Random~\cite{bergstra2011algorithms, li2020random}:} 
    It randomly samples graph models up to the budget and selects the one with the highest downstream-task performance.
    \item \textbf{S2. All-Rows-to-Nodes (\baseline) \cite{robinson2024relbench}:} This widely-used method includes all tables and foreign-key (FK) relationships in the graph, with all table rows modeled as nodes, as shown in Figure~\ref{fig:introexample}.
\end{itemize}

\textbf{Action-Based Search Algorithms.}
Action-based search algorithms explore and optimize graph models by iteratively modifying them using a predefined set of operations, referred to as \textit{actions}.

We consider the following four actions, which are designed to effectively span our graph model space: (1) \texttt{add\_fk\_edge:} adding an FK relationship between tables to the graph model, (2) \texttt{remove\_fk\_edge:} removing an existing FK relationship, (3) \texttt{convert\_row\_to\_edge:} changing the modeling of a table from \texttt{Row2Node} to \texttt{Row2Edge}, and (4) \texttt{convert\_edge\_to\_row:} changing the modeling of a table from \texttt{Row2Edge} to \texttt{Row2Node}.
We build the following six methods based on these actions:
\begin{itemize}[leftmargin=*]
    \item \textbf{S3. Greedy Forward Search (GF):} It starts from the target table and greedily repeats an action (except for \texttt{remove\_fk\_edge})
    that yields the greatest improvement to the current graph in terms of downstream-task performance. Recall that tables that are disconnected or distant from the target table are excluded from the graph model, as described in Section~\ref{subsec:searchspace}.
    
    \item \textbf{S4. Greedy Backward Search (GB):} It starts from the AR2N graph, which includes the entire RDB, and greedily repeats an action (except for 
    \texttt{add\_fk\_edge})
    that yields the greatest improvement to the current graph.
    \item \textbf{S5. Greedy Local Search (GL):} It starts from a random graph and greedily repeats an action of any type that provides the greatest improvement to the current graph.

    \item \textbf{S6. Evolutionary Algorithm (EA):} It applies evolutionary principles, including \textit{mutation} and \textit{selection}, to iteratively evolve graph models over generations.
    It randomly applies the predefined actions of any type to mutate the current graph and search for improved ones.
    Our implementation follows the regularized evolution strategy~\cite{real2019regularized}.
    \item \textbf{S7. Bayesian Optimization (BO):} It applies a Bayesian optimization algorithm, specifically BANANAS \cite{white2021bananas}, to the RDB-to-graph modeling task.
    This approach efficiently explores the graph model space by (a) modeling the function that maps actions to performance and (b) iteratively selecting actions of any type that are likely to improve the graph.
    \item \textbf{S8. Reinforcement Learning (RL):} It applies a controller based on recurrent neural networks~\cite{schuster1997bidirectional, hochreiter1997long} and trains it via a policy gradient descent approach~\cite{williams1992simple, baker2017designing}. The RL-based predictor learns to select actions of any type at each step of the RDB-to-graph modeling process to optimize downstream-task performance.
\end{itemize}
Further details of the action-based search algorithms are provided in Appendix~\ref{sec:appendix_bench}.

\textbf{LLM-Based Baselines.}
Inspired by AutoG~\cite{chen2025autog}, we developed the following two LLM-based baselines that generate action sequences for RDB-to-graph modeling using LLMs.
\begin{itemize}[leftmargin=*]
    \item \textbf{S9. LLM baseline (LLM):} It directly generates action sequences that specify how a given RDB is expressed as a graph based on the given prompts. At each step, an LLM is provided with a set of candidate actions and selects a sequence of actions expected to be most effective, based on its reasoning capabilities. The key distinction from AutoG lies in the simplified prompt design. Prompts irrelevant due to differences in action spaces are omitted.
    \item \textbf{S10. LLM-CoT baseline (LLM-CoT):} It applies Chain-of-Thought~\cite{wei2022chain} prompt design to encourage complex reasoning in the LLM baseline during the action selection step.
\end{itemize}
For fair comparisons with the above action-based search algorithms, both baselines operate over the same predefined set of action types.
\textbf{Claude Sonnet-3.5} is used as the backbone LLM, and details of the prompt designs are provided in Appendix~\ref{sec:appendix_prompt}.

\subsection{Benchmark Results}\label{subsec:benchmarkresults}

\begin{figure*}[t]
    \vspace{-3mm}
    \centering
    \begin{subfigure}[t]{0.585\linewidth}
        \centering
        \includegraphics[width=\linewidth]{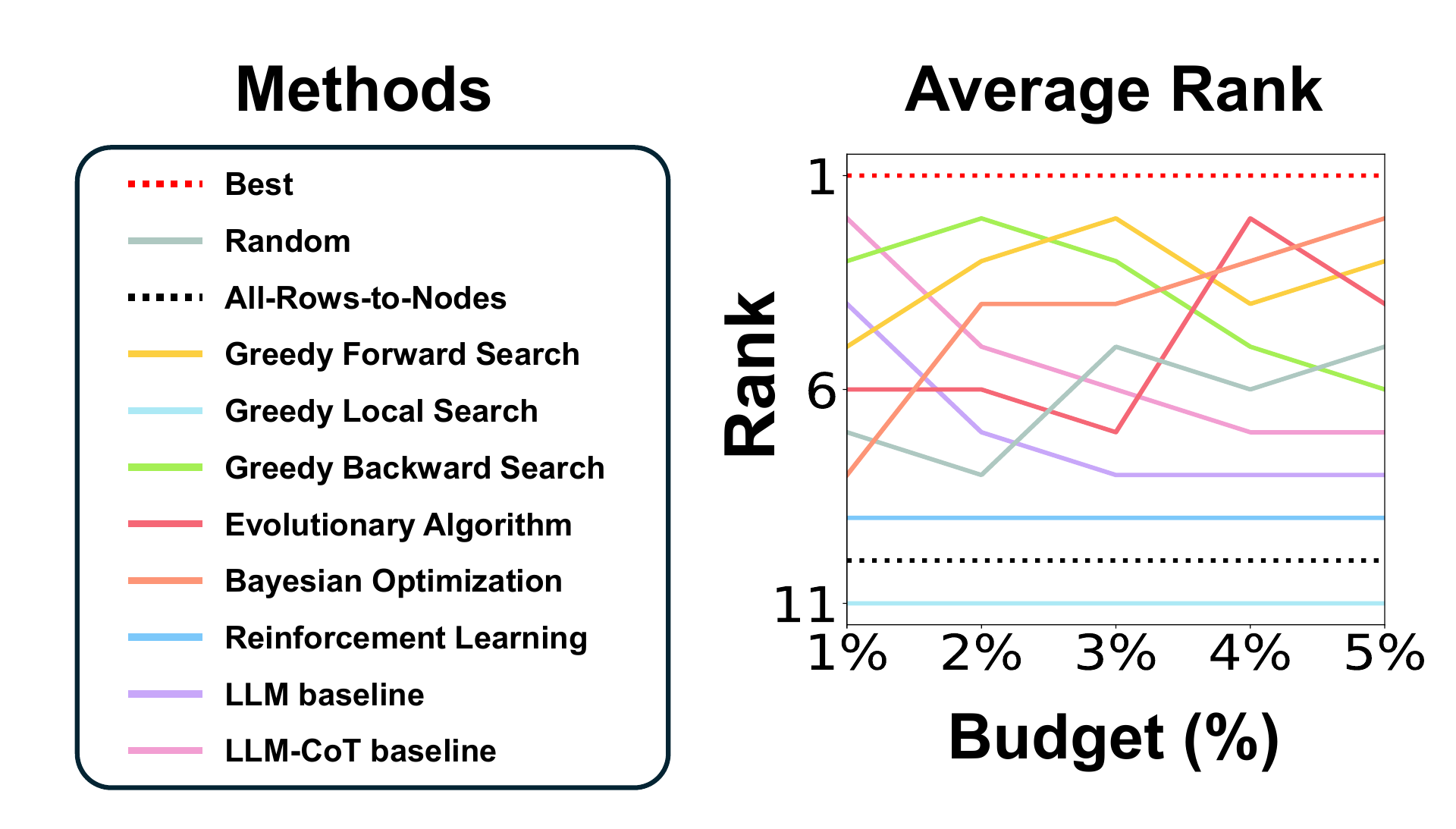}
        \caption{\textbf{Rank Averaged over Tasks}}
        \label{fig:benchmark_rank}
    \end{subfigure}
    \hfill
    \begin{subfigure}[t]{0.324\linewidth}
        \centering
        \includegraphics[width=\linewidth]{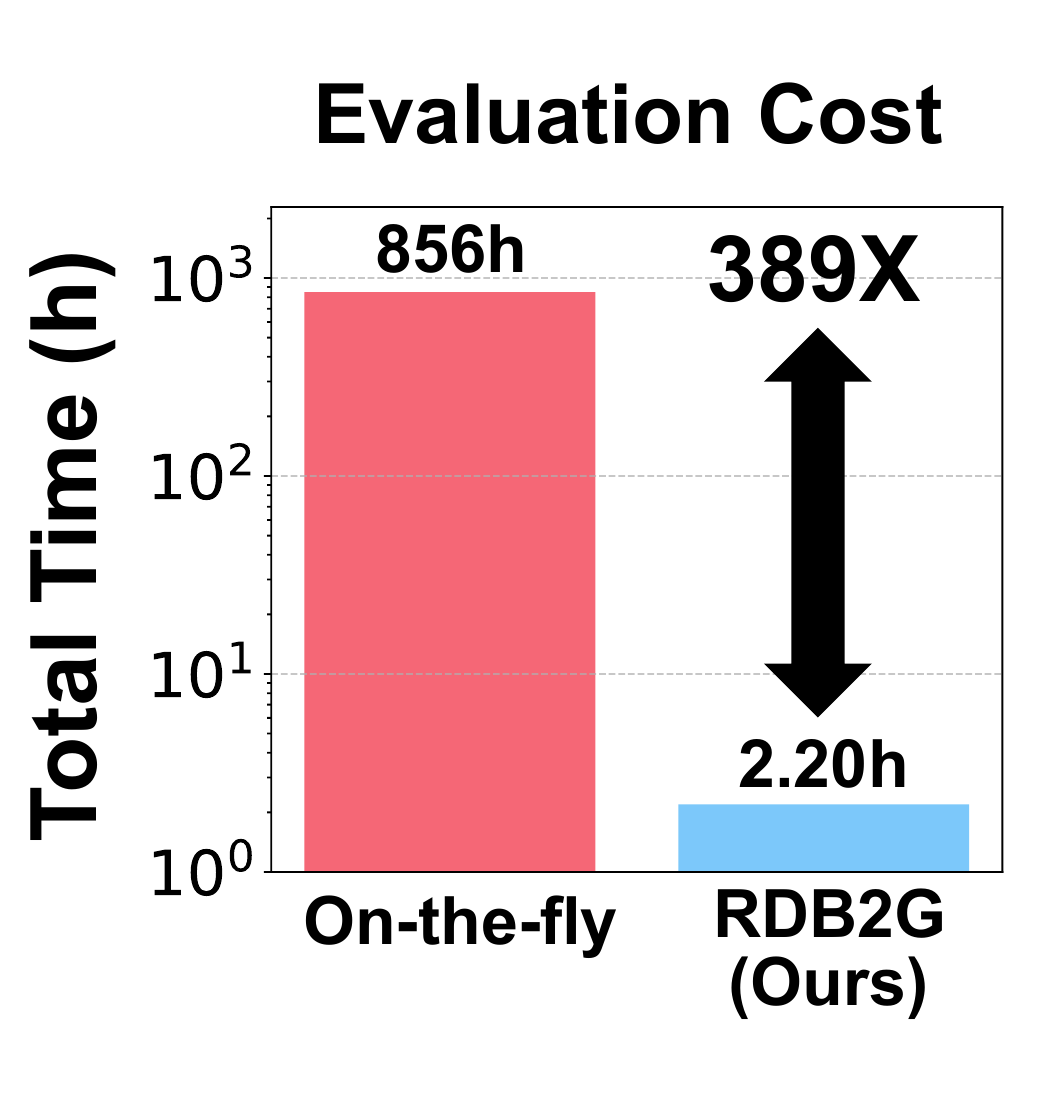}
        \caption{\textbf{Evaluation Cost}}
        \label{fig:benchmark_time}
    \end{subfigure}
    \caption{
        (a) Performance ranks of ten RDB-to-graph modeling methods, averaged across 12 predictive tasks.
        The ranks are computed under varying budget levels, corresponding to the number of graph models evaluated.
        (b) Comparison of evaluation costs (total elapsed time for all benchmark experiments) between two settings: on-the-fly and \ourdata. 
        \textbf{\ourdata speeds up benchmarking by 389$\times$} by eliminating on-the-fly graph model evaluation and associated GNN training.
    }
    \label{fig:benchresult}
\end{figure*}

For fair comparisons, we evaluate all baselines under a \textit{budget}, which limits the number of graph models whose effectiveness (e.g., ground-truth downstream-task performance) can be measured.
We track the effectiveness of resulting graph models as the budget increases up to 5\% of the total search space.
Full results, omitted here due to space constraints, are provided in Appendix~\ref{sec:appendix_benchres}.

Figure~\ref{fig:benchmark_rank} shows the ranks of the baselines, averaged over 12 predictive tasks from \ourdata and 10 independent runs per baseline (3 runs for the LLM-based method). 
In addition, Table~\ref{tab:benchfone} shows the details results on the \fone dataset.
Based on these results, we derive several key observations:
\begin{itemize}[leftmargin=*]
    \item Most baselines outperform \texttt{AR2N} modeling~\cite{robinson2024relbench} with minimal exploration, suggesting that actively searching for effective graph models is preferable to relying on a fixed modeling rule. 
    \item 
    Greedy methods and \texttt{Random} perform comparably to advanced approaches such as \texttt{Bayesian Optimization} and \texttt{Reinforcement Learning}.
    Especially, \texttt{Greedy Backward} performs best under small budgets, while \texttt{Bayesian Optimization} outperforms it as the budget increases. 
    Therefore, the choice between them can be guided by the available computational resources or time constraints.
    \item
    Value-based optimization algorithms, such as \texttt{Bayesian Optimization} and \texttt{Greedy Forward}, typically improve in effectiveness as performance feedback accumulates.
    However, under limited compute budgets, more complex algorithms, such as \texttt{Reinforcement Learning}, or algorithms affected by poor initialization, such as \texttt{Greedy Local}, often suffer from unstable exploration, resulting in poor performance.
    \item
     In the short run, the LLM-based approach demonstrated strong capabilities in rapidly improving performance through contextual reasoning and effective multi-turn planning.
    However, it struggled to devise effective long-term action plans, resulting in limited gains as the budget increased.
    CoT reasoning shows slight performance improvements over the LLM baseline in most cases, which provides potential for using LLMs in RDB-to-graph transformation.
\end{itemize}
The strong performance of simple methods like greedy approaches indicates significant room for improving RDB-to-graph modeling.

\vspace{-1mm}
\textbf{Efficiency Gain due to \ourdata:}
Our extensive benchmarking is made efficient by \ourdata, i.e., the precomputed performance metrics for graph models.
Without \ourdata, it is inevitable to evaluate graph models on the fly during the search process, requiring repeated training of graph neural networks.
As shown in Figure~\ref{fig:benchmark_time}, using \ourdata reduces evaluation time 389$\times$ from over 850 hours to just 2.20 hours. \ourdata equally benefits new RDB-to-graph modeling methods, fostering research in this direction.

\begin{table}[t]
  \centering
  \caption{Performance of ten RDB-to-graph modeling methods on the \fone dataset under varying budget levels.
  Refer to Appendix~\ref{app:benchresults} for results on other datasets.
  }
  \label{tab:benchfone}
  \setlength{\tabcolsep}{3pt}
  \vspace{1mm}
  \scalebox{0.73}{%
  \begin{tabular}{l|ccccc|ccccc|ccccc}
    \toprule
    \textbf{Task Name} &
    \multicolumn{5}{|c|}{\textbf{\driverDNF\ (AUC-ROC (\%) $\uparrow$)}} &
    \multicolumn{5}{c|}{\textbf{\driverTopThree\ (AUC-ROC (\%) $\uparrow$)}} &
    \multicolumn{5}{c}{\textbf{\driverPosition\ (MAE $\downarrow$)}} \\
    \midrule 
    \mr{2}{\textbf{Methods}} & \multicolumn{5}{c|}{\textbf{Budget (\%)}} &
    \multicolumn{5}{c|}{\textbf{Budget (\%)}} &
    \multicolumn{5}{c}{\textbf{Budget (\%)}} \\
    & \textbf{1\%} & \textbf{2\%} & \textbf{3\%} & \textbf{4\%} & \textbf{5\%}
    & \textbf{1\%} & \textbf{2\%} & \textbf{3\%} & \textbf{4\%} & \textbf{5\%}
    & \textbf{1\%} & \textbf{2\%} & \textbf{3\%} & \textbf{4\%} & \textbf{5\%} \\
    \midrule\midrule
    Best     & 74.557 & 74.557 & 74.557 & 74.557 & 74.557
             & 81.879 & 81.879 & 81.879 & 81.879 & 81.879
             & 3.8311 & 3.8311 & 3.8311 & 3.8311 & 3.8311 \\
             \midrule
    Random   & 73.225 & 73.450 & 73.592 & 73.745 & 73.755
             & 79.627 & 80.165 & 80.425 & 80.438 & 80.604
             & 3.8498 & 3.8435 & 3.8420 & 3.8405 & 3.8399 \\
    AR2N     & 73.140 & 73.140 & 73.140 & 73.140 & 73.140
             & 78.106 & 78.106 & 78.106 & 78.106 & 78.106
             & 3.9125 & 3.9125 & 3.9125 & 3.9125 & 3.9125 \\
    GF       & 73.461 & 74.557 & 74.557 & 74.557 & 74.557
             & 81.879 & 81.879 & 81.879 & 81.879 & 81.879
             & 3.8591 & 3.8409 & 3.8409 & 3.8409 & 3.8409 \\
    GB       & 73.254 & 74.040 & 74.394 & 74.394 & 74.394
             & 80.148 & 80.558 & 80.558 & 80.558 & 80.558
             & 3.8437 & 3.8437 & 3.8437 & 3.8437 & 3.8437 \\
    GL       & 73.023 & 73.645 & 73.726 & 73.775 & 73.775
             & 79.172 & 79.292 & 79.299 & 79.299 & 79.299
             & 3.8649 & 3.8512 & 3.8500 & 3.8500 & 3.8500 \\
    EA       & 73.082 & 73.393 & 73.590 & 73.786 & 73.929
             & 79.153 & 79.774 & 80.116 & 80.273 & 80.549
             & 3.8470 & 3.8451 & 3.8423 & 3.8394 & 3.8394 \\
    BO       & 73.328 & 73.624 & 73.837 & 73.919 & 74.070
             & 79.424 & 79.799 & 80.127 & 80.152 & 80.348
             & 3.8520 & 3.8404 & 3.8399 & 3.8394 & 3.8388 \\
    RL       & 73.332 & 73.526 & 73.602 & 73.657 & 73.986
             & 79.043 & 79.385 & 79.439 & 79.906 & 80.084
             & 3.8542 & 3.8449 & 3.8417 & 3.8411 & 3.8411 \\
    LLM      & 73.857 & 73.857 & 73.857 & 73.857 & 73.857
             & 80.337 & 80.536 & 80.536 & 80.607 & 80.607
             & 3.8534 & 3.8534 & 3.8500 & 3.8500 & 3.8500 \\
    LLM-CoT  & 73.338 & 73.574 & 73.882 & 73.882 & 73.882
             & 78.660 & 78.693 & 78.693 & 78.817 & 78.817
             & 3.8437 & 3.8437 & 3.8437 & 3.8437 & 3.8437 \\
    \bottomrule
  \end{tabular}
  }% end resizebox
\end{table}

\section{Conclusions}
\label{sec:conclusion}
This study presents \ourdata, the first benchmark designed to evaluate automatic graph modeling methods for relational databases (RDBs).
For \ourdata, we precomputed 50k graph models derived from 5 real-world RDBs and 12 predictive tasks, along with their associated downstream performance, runtime, and parameter size.
Analysis of the \ourdata datasets reveals that more data or additional tables do not necessarily improve predictive performance. Instead, selectively using fewer, relevant tables often leads to better results with improved efficiency. Moreover, the best-performing graph models vary by task, emphasizing the need for intelligent modeling strategies that account for the characteristics of both RDBs and predictive tasks. Moreover, our extensive benchmark of ten RDB-to-graph modeling strategies reveals substantial room for improvement in this domain, with LLM-based reasoning showing promising potential, despite its current limitations. \ourdata, whose datasets and code are publicly available at \url{https://github.com/chlehdwon/RDB2G-Bench}, significantly accelerates the evaluation of diverse RDB-to-graph modeling strategies, by up to 389$\times$, thereby facilitating the development of more advanced techniques.

%%%%%%%%%%%%%%%%%%%%%%%%%%%%%%%%%%%%%%%%%%%%%%%%%%%%%%%%%%%%

\section*{Acknowledgements}
This work was partly supported by the National Research Foundation of Korea (NRF) grant funded by the Korea government (MSIT) (No. RS-2024-00406985, 20\%).
This work was partly supported  by Institute of Information \& Communications Technology Planning \& Evaluation (IITP) grant funded by the Korea government (MSIT) (No. RS-2024-00438638, EntireDB2AI: Foundations and Software for Comprehensive Deep Representation Learning and Prediction on Entire Relational Databases, 50\%)
(No. RS-2024-00457882, AI Research Hub Project, 10\%)
(RS-2025-02653113, High-Performance Research AI Computing Infrastructure Support at the 2 PFLOPS Scale, 10\%) (RS-2019-II190075, Artificial Intelligence Graduate School Program (KAIST), 10\%).

\newpage

\bibliographystyle{abbrvnat}
\bibliography{neurips_2025}

\clearpage

\section{Checklist}
%Optionally include supplemental material (complete proofs, additional experiments and plots) in appendix.
%All such materials \textbf{SHOULD be included in the main submission.}

%%%%%%%%%%%%%%%%%%%%%%%%%%%%%%%%%%%%%%%%%%%%%%%%%%%%%%%%%%%%

\section*{NeurIPS Paper Checklist}

\begin{enumerate}

\item {\bf Claims}
    \item[] Question: Do the main claims made in the abstract and introduction accurately reflect the paper's contributions and scope?
    \item[] Answer: \answerYes{} % Replace by \answerYes{}, \answerNo{}, or \answerNA{}.
    \item[] Justification: We reflect all the contributions and scope in the abstract and introduction(Section ~\ref{sec:introduction}). In detail, we provide 3 contributions while describing the main claim of our paper.
    % \item[] Guidelines:
    % \begin{itemize}
    %     \item The answer NA means that the abstract and introduction do not include the claims made in the paper.
    %     \item The abstract and/or introduction should clearly state the claims made, including the contributions made in the paper and important assumptions and limitations. A No or NA answer to this question will not be perceived well by the reviewers. 
    %     \item The claims made should match theoretical and experimental results, and reflect how much the results can be expected to generalize to other settings. 
    %     \item It is fine to include aspirational goals as motivation as long as it is clear that these goals are not attained by the paper. 
    % \end{itemize}

\item {\bf Limitations}
    \item[] Question: Does the paper discuss the limitations of the work performed by the authors?
    \item[] Answer: \answerYes{} % Replace by \answerYes{}, \answerNo{}, or \answerNA{}.
    \item[] Justification: We describe our limitations in Appendix~\ref{app:limitations}.

\item {\bf Theory assumptions and proofs}
    \item[] Question: For each theoretical result, does the paper provide the full set of assumptions and a complete (and correct) proof?
    \item[] Answer: \answerNA{} % Replace by \answerYes{}, \answerNo{}, or \answerNA{}.
    \item[] Justification: Our paper doesn't include any theoretical results.
    % \item[] Guidelines:
    % \begin{itemize}
    %     \item The answer NA means that the paper does not include theoretical results. 
    %     \item All the theorems, formulas, and proofs in the paper should be numbered and cross-referenced.
    %     \item All assumptions should be clearly stated or referenced in the statement of any theorems.
    %     \item The proofs can either appear in the main paper or the supplemental material, but if they appear in the supplemental material, the authors are encouraged to provide a short proof sketch to provide intuition. 
    %     \item Inversely, any informal proof provided in the core of the paper should be complemented by formal proofs provided in appendix or supplemental material.
    %     \item Theorems and Lemmas that the proof relies upon should be properly referenced. 
    % \end{itemize}

    \item {\bf Experimental result reproducibility}
    \item[] Question: Does the paper fully disclose all the information needed to reproduce the main experimental results of the paper to the extent that it affects the main claims and/or conclusions of the paper (regardless of whether the code and data are provided or not)?
    \item[] Answer: \answerYes{} % Replace by \answerYes{}, \answerNo{}, or \answerNA{}.
    \item[] Justification: Our results are reproducible because all experiment results are computed from \ourdata, which we provide in this paper. We also release our datasets and codes in public (Github, Huggingface).

\item {\bf Open access to data and code}
    \item[] Question: Does the paper provide open access to the data and code, with sufficient instructions to faithfully reproduce the main experimental results, as described in supplemental material?
    \item[] Answer: \answerYes{} % Replace by \answerYes{}, \answerNo{}, or \answerNA{}.
    \item[] Justification: We provide all the codes and datasets through \ourdata. We also construct documents to faithfully reproduce the main experimental results. You can check all the code and data from the following URLs:
    \begin{itemize}
        \item \textbf{Dataset: } \huggingface
        \item \textbf{Code: } \github
    \end{itemize}
    % \item[] Guidelines:
    % \begin{itemize}
    %     \item The answer NA means that paper does not include experiments requiring code.
    %     \item Please see the NeurIPS code and data submission guidelines (\url{https://nips.cc/public/guides/CodeSubmissionPolicy}) for more details.
    %     \item While we encourage the release of code and data, we understand that this might not be possible, so “No” is an acceptable answer. Papers cannot be rejected simply for not including code, unless this is central to the contribution (e.g., for a new open-source benchmark).
    %     \item The instructions should contain the exact command and environment needed to run to reproduce the results. See the NeurIPS code and data submission guidelines (\url{https://nips.cc/public/guides/CodeSubmissionPolicy}) for more details.
    %     \item The authors should provide instructions on data access and preparation, including how to access the raw data, preprocessed data, intermediate data, and generated data, etc.
    %     \item The authors should provide scripts to reproduce all experimental results for the new proposed method and baselines. If only a subset of experiments are reproducible, they should state which ones are omitted from the script and why.
    %     \item At submission time, to preserve anonymity, the authors should release anonymized versions (if applicable).
    %     \item Providing as much information as possible in supplemental material (appended to the paper) is recommended, but including URLs to data and code is permitted.
    % \end{itemize}

\item {\bf Experimental setting/details}
    \item[] Question: Does the paper specify all the training and test details (e.g., data splits, hyperparameters, how they were chosen, type of optimizer, etc.) necessary to understand the results?
    \item[] Answer: \answerYes{} % Replace by \answerYes{}, \answerNo{}, or \answerNA{}.
    \item[] Justification: We mentioned our hyperparameter settings in Section~\ref{subsec:datasettings} and Section~\ref{subsec:benchmarksettings}. Also all the details are described in Appendix~\ref{app:datasethyperparams} and Appendix~\ref{app:baselinedetail}.
    % \item[] Guidelines:
    % \begin{itemize}
    %     \item The answer NA means that the paper does not include experiments.
    %     \item The experimental setting should be presented in the core of the paper to a level of detail that is necessary to appreciate the results and make sense of them.
    %     \item The full details can be provided either with the code, in appendix, or as supplemental material.
    % \end{itemize}

\item {\bf Experiment statistical significance}
    \item[] Question: Does the paper report error bars suitably and correctly defined or other appropriate information about the statistical significance of the experiments?
    \item[] Answer: \answerYes{} % Replace by \answerYes{}, \answerNo{}, or \answerNA{}.
    \item[] Justification: We report standard deviations in Figure~\ref{fig:introexample} and error bars in Figure~\ref{fig:datasetsummary}, and leverage box plots in Figure~\ref{fig:row2nemain} to justify the statistical significance of our experiments.
    % \item[] Guidelines:
    % \begin{itemize}
    %     \item The answer NA means that the paper does not include experiments.
    %     \item The authors should answer "Yes" if the results are accompanied by error bars, confidence intervals, or statistical significance tests, at least for the experiments that support the main claims of the paper.
    %     \item The factors of variability that the error bars are capturing should be clearly stated (for example, train/test split, initialization, random drawing of some parameter, or overall run with given experimental conditions).
    %     \item The method for calculating the error bars should be explained (closed form formula, call to a library function, bootstrap, etc.)
    %     \item The assumptions made should be given (e.g., Normally distributed errors).
    %     \item It should be clear whether the error bar is the standard deviation or the standard error of the mean.
    %     \item It is OK to report 1-sigma error bars, but one should state it. The authors should preferably report a 2-sigma error bar than state that they have a 96\% CI, if the hypothesis of Normality of errors is not verified.
    %     \item For asymmetric distributions, the authors should be careful not to show in tables or figures symmetric error bars that would yield results that are out of range (e.g. negative error rates).
    %     \item If error bars are reported in tables or plots, The authors should explain in the text how they were calculated and reference the corresponding figures or tables in the text.
    % \end{itemize}

\item {\bf Experiments compute resources}
    \item[] Question: For each experiment, does the paper provide sufficient information on the computer resources (type of compute workers, memory, time of execution) needed to reproduce the experiments?
    \item[] Answer: \answerYes{} % Replace by \answerYes{}, \answerNo{}, or \answerNA{}.
    \item[] Justification: We describe all the computer resources and time of execution in Section~\ref{subsec:datasettings} (See \textbf{Machines.}).
    % \item[] Guidelines:
    % \begin{itemize}
    %     \item The answer NA means that the paper does not include experiments.
    %     \item The paper should indicate the type of compute workers CPU or GPU, internal cluster, or cloud provider, including relevant memory and storage.
    %     \item The paper should provide the amount of compute required for each of the individual experimental runs as well as estimate the total compute. 
    %     \item The paper should disclose whether the full research project required more compute than the experiments reported in the paper (e.g., preliminary or failed experiments that didn't make it into the paper). 
    % \end{itemize}
    
\item {\bf Code of ethics}
    \item[] Question: Does the research conducted in the paper conform, in every respect, with the NeurIPS Code of Ethics \url{https://neurips.cc/public/EthicsGuidelines}?
    \item[] Answer: \answerYes{} % Replace by \answerYes{}, \answerNo{}, or \answerNA{}.
    \item[] Justification: We strictly follow the NeurIPS Code of Ethics, particularly the source of our dataset, which RelBench has already anonymized all the data. Also, we don't contain any privacy information in our datasets and codes.
    % \item[] Guidelines:
    % \begin{itemize}
    %     \item The answer NA means that the authors have not reviewed the NeurIPS Code of Ethics.
    %     \item If the authors answer No, they should explain the special circumstances that require a deviation from the Code of Ethics.
    %     \item The authors should make sure to preserve anonymity (e.g., if there is a special consideration due to laws or regulations in their jurisdiction).
    % \end{itemize}

\item {\bf Broader impacts}
    \item[] Question: Does the paper discuss both potential positive societal impacts and negative societal impacts of the work performed?
    \item[] Answer: \answerYes{} % Replace by \answerYes{}, \answerNo{}, or \answerNA{}.
    \item[] Justification: We discuss various broader impacts of our paper in Appendix~\ref{app:broaderimpact}.

\item {\bf Safeguards}
    \item[] Question: Does the paper describe safeguards that have been put in place for responsible release of data or models that have a high risk for misuse (e.g., pretrained language models, image generators, or scraped datasets)?
    \item[] Answer: \answerNA{} % Replace by \answerYes{}, \answerNo{}, or \answerNA{}.
    \item[] Justification: We believe we do not have risks of misusing.
    % \item[] Guidelines:
    % \begin{itemize}
    %     \item The answer NA means that the paper poses no such risks.
    %     \item Released models that have a high risk for misuse or dual-use should be released with necessary safeguards to allow for controlled use of the model, for example by requiring that users adhere to usage guidelines or restrictions to access the model or implementing safety filters. 
    %     \item Datasets that have been scraped from the Internet could pose safety risks. The authors should describe how they avoided releasing unsafe images.
    %     \item We recognize that providing effective safeguards is challenging, and many papers do not require this, but we encourage authors to take this into account and make a best faith effort.
    % \end{itemize}

\item {\bf Licenses for existing assets}
    \item[] Question: Are the creators or original owners of assets (e.g., code, data, models), used in the paper, properly credited and are the license and terms of use explicitly mentioned and properly respected?
    \item[] Answer: \answerYes{} % Replace by \answerYes{}, \answerNo{}, or \answerNA{}.
    \item[] Justification: We used the datasets and models of RelBench~\cite{robinson2024relbench} implementation, which mentioned several times in our paper. Also, there is no license violation in our paper.
    % \item[] Guidelines:
    % \begin{itemize}
    %     \item The answer NA means that the paper does not use existing assets.
    %     \item The authors should cite the original paper that produced the code package or dataset.
    %     \item The authors should state which version of the asset is used and, if possible, include a URL.
    %     \item The name of the license (e.g., CC-BY 4.0) should be included for each asset.
    %     \item For scraped data from a particular source (e.g., website), the copyright and terms of service of that source should be provided.
    %     \item If assets are released, the license, copyright information, and terms of use in the package should be provided. For popular datasets, \url{paperswithcode.com/datasets} has curated licenses for some datasets. Their licensing guide can help determine the license of a dataset.
    %     \item For existing datasets that are re-packaged, both the original license and the license of the derived asset (if it has changed) should be provided.
    %     \item If this information is not available online, the authors are encouraged to reach out to the asset's creators.
    % \end{itemize}

\item {\bf New assets}
    \item[] Question: Are new assets introduced in the paper well documented and is the documentation provided alongside the assets?
    \item[] Answer: \answerYes{} % Replace by \answerYes{}, \answerNo{}, or \answerNA{}.
    \item[] Justification: We release \ourdata which includes datasets and codes. We provide the documentation to help users that can be found in our GitHub link: \github, and Hugging Face link: \huggingface.
    % \item[] Guidelines:
    % \begin{itemize}
    %     \item The answer NA means that the paper does not release new assets.
    %     \item Researchers should communicate the details of the dataset/code/model as part of their submissions via structured templates. This includes details about training, license, limitations, etc. 
    %     \item The paper should discuss whether and how consent was obtained from people whose asset is used.
    %     \item At submission time, remember to anonymize your assets (if applicable). You can either create an anonymized URL or include an anonymized zip file.
    % \end{itemize}

\item {\bf Crowdsourcing and research with human subjects}
    \item[] Question: For crowdsourcing experiments and research with human subjects, does the paper include the full text of instructions given to participants and screenshots, if applicable, as well as details about compensation (if any)? 
    \item[] Answer: \answerNA{} % Replace by \answerYes{}, \answerNo{}, or \answerNA{}.
    \item[] Justification: We don't use any crowdsourcing or human-related tasks in our paper.
    % \item[] Guidelines:
    % \begin{itemize}
    %     \item The answer NA means that the paper does not involve crowdsourcing nor research with human subjects.
    %     \item Including this information in the supplemental material is fine, but if the main contribution of the paper involves human subjects, then as much detail as possible should be included in the main paper. 
    %     \item According to the NeurIPS Code of Ethics, workers involved in data collection, curation, or other labor should be paid at least the minimum wage in the country of the data collector. 
    % \end{itemize}

\item {\bf Institutional review board (IRB) approvals or equivalent for research with human subjects}
    \item[] Question: Does the paper describe potential risks incurred by study participants, whether such risks were disclosed to the subjects, and whether Institutional Review Board (IRB) approvals (or an equivalent approval/review based on the requirements of your country or institution) were obtained?
    \item[] Answer: \answerNA{} % Replace by \answerYes{}, \answerNo{}, or \answerNA{}.
    \item[] Justification: We do not utilize any crowdsourcing or human-related tasks.
    % \item[] Guidelines:
    % \begin{itemize}
    %     \item The answer NA means that the paper does not involve crowdsourcing nor research with human subjects.
    %     \item Depending on the country in which research is conducted, IRB approval (or equivalent) may be required for any human subjects research. If you obtained IRB approval, you should clearly state this in the paper. 
    %     \item We recognize that the procedures for this may vary significantly between institutions and locations, and we expect authors to adhere to the NeurIPS Code of Ethics and the guidelines for their institution. 
    %     \item For initial submissions, do not include any information that would break anonymity (if applicable), such as the institution conducting the review.
    % \end{itemize}

\item {\bf Declaration of LLM usage}
    \item[] Question: Does the paper describe the usage of LLMs if it is an important, original, or non-standard component of the core methods in this research? Note that if the LLM is used only for writing, editing, or formatting purposes and does not impact the core methodology, scientific rigorousness, or originality of the research, declaration is not required.
    %this research? 
    \item[] Answer: \answerYes{} % Replace by \answerYes{}, \answerNo{}, or \answerNA{}.
    \item[] Justification: We implemented LLM-based method as a benchmark baseline, which motivated by AutoG~\cite{chen2025autog} paper. The differences between AutoG are described in Section~\ref{subsec:benchmarksettings}, and all the prompts that we used are in Appendix~\ref{sec:appendix_prompt}.
    % \item[] Guidelines:
    % \begin{itemize}
    %     \item The answer NA means that the core method development in this research does not involve LLMs as any important, original, or non-standard components.
    %     \item Please refer to our LLM policy (\url{https://neurips.cc/Conferences/2025/LLM}) for what should or should not be described.
    % \end{itemize}

\end{enumerate}

\label{sec:checklist}

\newpage 
\appendix
\section{Dataset Detail of the \ourdata}
\label{sec:appendix_dataset}
\begin{table}[ht]
  \centering
  \caption{{Statistics of \relbench.}  }
  \label{tab:stats}
  \setlength{\tabcolsep}{4pt}
  \scalebox{0.85}{
  \begin{tabular}{lll|ccc|cc}
    \toprule
      \multirow{2}[2]{*}{\textbf{RDB}} & \multirow{2}[2]{*}{\textbf{Task Name}} &\multirow{2}[2]{*}{\textbf{Type}} & \mc{3}{c|}{\textbf{\#Rows of Training Table}} & \multirow{2}[2]{*}{\textbf{\makecell{\#Unique\\Entities}}} & \multirow{2}[2]{*}{\textbf{\makecell{\%Train/Test\\Entity Overlap}}}  \\
      \cmidrule{4-6}
      & & & \textbf{Train} & \textbf{Validation} & \textbf{Test} & & \\
    \midrule
    \midrule
    \mr{4}{\avito}
    & \userClick & classification & 59,454 & 21,183 & 47,996 & 66,449 & 45.3 \\
    & \userVisit & classification & 86,619 & 29,979 & 36,129 & 63,405 & 64.6 \\
    & \adsCTR & regression & 5,100 & 1,766 & 1,816 & 4,997 & 59.8 \\
    & \userAdVisit & recommendation & 86,616 & 29,979 & 36,129 & 63,402 & 64.6 \\
    \midrule
    \mr{3}{\event}
    & \userRepeat & classification & 3,842 & 268 & 246 &1,514 & 11.5\\
    & \userIgnore & classification & 19,239 & 4,185 & 4,010 &9,799 & 21.1\\
    & \userAttendance & regression & 19,261 & 2,014 & 2,006 &9,694 & 14.6\\
    \midrule
    \mr{3}{\fone}
      & \driverDNF & classification & 11,411 & 566 & 702 & 821 & 50.0  \\
      & \driverTopThree & classification  & 1,353 & 588 & 726 & 134 & 50.0  \\
      & \driverPosition & regression & 7,453 & 499 & 760 & 826 & 44.6  \\
    \midrule
    \mr{1}{\stackex}
      & \postPostLinked & recommendation  & 5,855 & 226 & 258 & 5,924 & 8.5  \\
    \midrule
    \mr{1}{\trials}
      & \studyOutcome & classification  & 11,994 & 960 & 825 & 13,779 & 0.0  \\
   \bottomrule
  \end{tabular}
  }
\end{table}

\subsection{\relbench Detail}
\label{app:relbenchdataset}
Our research involves five datasets and twelve tasks from \relbench~\cite{robinson2024relbench}, each representing diverse domains and varying scales. A detailed description of each dataset and task is provided below. For more details, please refer to the URLs. Additional statistics are presented in Table \ref{tab:stats}.

\xhdr{\avito~\protect\footnotemark}
Avito is a leading online advertisement platform, providing a marketplace for users to buy and sell a wide variety of products and services, including real estate, vehicles, jobs, and goods. The Avito Context Ad Clicks dataset on Kaggle is part of a competition aimed at predicting whether an ad will be clicked based on contextual information. This dataset includes user searches, ad attributes, and other related data to help build predictive GNNs.

\footnotetext{\url{https://relbench.stanford.edu/datasets/rel-avito/}}
\begin{enumerate}[leftmargin=*]
    \item \userVisit: Predict whether each customer will visit more than one Ad in the next 4 days.
    \item \userClick: Predict whether each customer will click on more than one Ads in the next 4 days.
    \item \adsCTR: Assuming the Ad will be clicked in the next 4 days, predict the Click-Through-Rate (CTR) for each Ad.
    \item \userAdVisit: Predict the list of ads a user will visit in the next 4 days.
\end{enumerate}

\xhdr{\event~\protect\footnotemark} The Event Recommendation database is obtained from user data on a mobile app called Hangtime. This app allows users to keep track of their friends' social plans. The database contains data on user actions, event metadata, and demographic information, as well as users' social relations, which captures how social relations can affect user behavior.

\footnotetext{\url{https://relbench.stanford.edu/datasets/rel-event/}}
\begin{enumerate}[leftmargin=*]
    \item \userRepeat: Predict whether a user will attend an event(by responding yes or maybe) in the next 7 days if they have already attended an event in the last 14 days.
    \item \userIgnore: Predict whether a user will ignore more than 2 event invitations in the next 7 days
    \item \userAttendance: Predict how many events each user will respond yes or maybe in the next seven days.
\end{enumerate}

\xhdr{\fone~\protect\footnotemark} The F1 database tracks all-time Formula 1 racing data and statistics since 1950. It provides detailed information for various stakeholders including drivers, constructors, engine manufacturers, and tyre manufacturers. Highlights include data on all circuits (e.g. geographical details), and full historical data from every season. This includes overall standings, race results, and more specific data like practice sessions, qualifying positions, sprints, and pit stops.

\footnotetext{\url{https://relbench.stanford.edu/datasets/rel-f1/}}
\begin{enumerate}
    \item \driverDNF: For each driver predict the if they will DNF (did not finish) a race in the next 1 month.
    \item \driverTopThree: For each driver predict if they will qualify in the top-3 for a race in the next 1 month.
    \item \driverPosition: Predict the average finishing position of each driver all races in the next 2 months.
\end{enumerate}

\xhdr{\stackex~\protect\footnotemark} Stack Exchange is a network of question-and-answer websites on topics in diverse fields, each site covering a specific topic, where questions, answers, and users are subject to a reputation award process. The reputation system allows the sites to be self-moderating.

\footnotetext{\url{https://relbench.stanford.edu/datasets/rel-stack/}}
\begin{enumerate}
    \item \postPostLinked: Predict a list of existing posts that users will link a given post to in the next two years.
\end{enumerate}

\xhdr{\trials~\protect\footnotemark} The clinical trial database is curated from AACT initiative, which consolidates all protocol and results data from studies registered on ClinicalTrials.gov. It offers extensive information about clinical trials, including study designs, participant demographics, intervention details, and outcomes. It is an important resource for health research, policy making, and therapeutic development.

\footnotetext{\url{https://relbench.stanford.edu/datasets/rel-trial/}}
\begin{enumerate}
    \item \studyOutcome: Predict if the trials will achieve its primary outcome.
\end{enumerate}

\subsection{GNN Implementation}
\label{app:relbenchmodel}

In this section, we demonstrate the implementations of GraphSAGE, GIN, and GPS, including message passing and edge feature extensions for \texttt{Row2Edge} modeling. All implementations are based on the PyTorch Geometric (PyG)~\cite{fey2019fast} library.

\textbf{GraphSAGE Implementation.} The standard GraphSAGE~\cite{hamilton2017inductive} message passing with \textbf{sum aggregation} is defined as:

\begin{equation}\label{eq:sageconv}
\mathbf{h}_i^{(k)} = \sigma\left( 
\mathbf{W}^{(k)} \cdot 
\sum_{j \in \mathcal{N}(i)} \mathbf{h}_j^{(k-1)}
+ \mathbf{W}_r^{(k)} \cdot \mathbf{h}_i^{(k-1)} 
\right)
\end{equation}

Where, $\mathbf{h}_i^{(k)}$ is embedding of node $i$ at layer $k$, $\mathcal{N}(i)$ represents set of neighboring nodes of $i$, and $\sigma$ denotes non-linear activation function (e.g., ReLU).

To extend this formulation to incorporate edge features, $\mathbf{e}_{j \rightarrow i}$ for \texttt{Row2Edge}~\cite{wang20244dbinfer}, we concatenate them with neighbor features. Then, \texttt{Row2Edge} message passing becomes as follows (the modified part is highlighted in \red{red}):

\begin{equation}\label{eq:sageedgeconv}
\mathbf{h}_i^{(k)} = \sigma\left( 
\mathbf{W}^{(k)} \cdot 
\sum_{j \in \mathcal{N}(i)} 
\textcolor{red}{\left[ \mathbf{h}_j^{(k-1)} \, \| \, \mathbf{e}_{j \rightarrow i} \right]}
+ \mathbf{W}_r^{(k)} \cdot \mathbf{h}_i^{(k-1)} 
\right)
\end{equation}

GraphSAGE message passing with \textbf{mean aggregation}, which used in Section~\ref{subsec:modelgeneralization} is defined as:
\begin{equation}\label{eq:sageconvmean}
\mathbf{h}_i^{(k)} = \sigma\left( 
\mathbf{W}^{(k)} \cdot 
\frac{1}{N}\sum_{j \in \mathcal{N}(i)} \mathbf{h}_j^{(k-1)}
+ \mathbf{W}_r^{(k)} \cdot \mathbf{h}_i^{(k-1)} 
\right)
\end{equation}

Where $N$ is defined as the number of neighbors.

\textbf{GIN Implementation.} Graph Isomorphism Network (GIN)~\cite{xu2019powerful} message passing is defined as:

\begin{equation}\label{eq:ginconv}
\mathbf{h}_i^{(k)} = \mathrm{MLP}^{(k)} \left( 
(1 + \epsilon^{(k)}) \cdot \mathbf{h}_i^{(k-1)} + 
\sum_{j \in \mathcal{N}(i)} \mathbf{h}_j^{(k-1)} 
\right)
\end{equation}

Here, $\mathbf{h}_i^{(k)}$ denotes the embedding of node $i$ at layer $k$, $\epsilon^{(k)}$ is a trainable scalar parameter, and $\mathrm{MLP}^{(k)}$ is a Multi-Layer Perceptron (MLP) network used to transform aggregated messages. In our implementation, we adopt the following form:

\begin{equation}\label{eq:ginmlp}
\mathrm{MLP}^{(k)}(\cdot) = \mathrm{Linear}(d, 2d) \rightarrow \mathrm{ReLU} \rightarrow \mathrm{Linear}(2d, d)
\end{equation}

where $d$ is the hidden dimensionality of node features.

To incorporate edge features $\mathbf{e}_{j \rightarrow i}$, we adopt the GINE~\cite{hu2020strategies} formulation. Specifically, neighbor embeddings are modulated with their corresponding edge features before aggregation, as follows (the modified part is highlighted in \red{red}):

\begin{equation}\label{eq:gineconv}
\mathbf{h}_i^{(k)} = \mathrm{MLP}^{(k)} \left(
(1 + \epsilon^{(k)}) \cdot \mathbf{h}_i^{(k-1)} + 
\sum_{j \in \mathcal{N}(i)} \textcolor{red}{\mathrm{ReLU}(\mathbf{h}_j^{(k-1)} + \mathbf{e}_{j \rightarrow i})}
\right)
\end{equation}

\textbf{GPS Implementation.} The General, Powerful, Scalable Graph Transformer (GPS)~\cite{rampavsek2022recipe} integrates local message passing with global attention, enabling expressive and scalable graph learning. GPS combines two components at each layer: (1) local neighborhood aggregation, and (2) global message exchange via Transformer-style self-attention.

Formally, for each layer $k$, the message passing is performed in two steps:
\begin{equation}
\mathbf{h}_i^{(k)} = \mathrm{LocalMP}^{(k)}\left(\{\mathbf{h}_j^{(k-1)} : j \in \mathcal{N}(i)\}\right) + 
\mathrm{GlobalAttn}^{(k)}\left(\{\mathbf{h}_j^{(k-1)} : j \in \mathcal{V}\}\right)
\end{equation}

where $\mathbf{h}_i^{(k)}$ is the embedding of node $i$ at layer $k$, $\mathcal{N}(i)$ denotes the set of neighboring nodes of $i$, and $\mathcal{V}$ is the set of all nodes in the graph. 

In our implementation, we employ Equation~(\ref{eq:sageconv}) and Equation~(\ref{eq:sageedgeconv}) for the $\mathrm{LocalMP}^{(k)}$ of \texttt{Row2Node} and \texttt{Row2Edge} modeling, respectively, to naturally extend our settings to transformer-based architectures. To fit the GPS within a single GPU, we reduce the number of sampled neighbors from 128 to 32.

\subsection{Hyperparameter Details}
\label{app:datasethyperparams}

Table~\ref{tab:hyperparam} shows our learning rate choices, which are only tuned for each task. For the other hyperparameter, all tasks used two-layer GNNs with a batch size of 512, 128 dimensions, and the Adam~\cite{kingma2014adam} optimizer.

Specifically, neighbor sampling counts for subgraph extraction are set to 128 when all tables are represented as nodes, and 12 when some tables are represented as edges.
A lower sampling count is employed when tables are modeled as edges, since the two-layer GNN effectively captures paths longer than two hops, resulting in a significantly larger graph scale that cannot be processed by a single GPU.

\begin{table}[ht]  
\centering
\caption{{The learning rate settings tuned for each task.}}
  \label{tab:hyperparam}
  \setlength{\tabcolsep}{4pt}
  \small
    \begin{tabular}{lllc}
    \toprule
    \textbf{RDB} & \textbf{Task name} & \textbf{Type} & \textbf{Learning Rate} \\
    \midrule
    \midrule
    \multirow{4}{*}{\texttt{rel-avito}} 
    & user-clicks & classification & 0.001 \\
    & user-visits & classification & 0.001 \\
    & ad-ctr & regression & 0.0005 \\
    & user-ad-visit & recommendation & 0.001 \\
    \midrule
    \multirow{3}{*}{\texttt{rel-event}} 
    & user-repeat & classification & 0.005 \\
    & user-ignore & classification & 0.005 \\
    & user-attendance & regression & 0.005 \\
    \midrule
    \multirow{3}{*}{\texttt{rel-f1}} 
    & driver-dnf & classification & 0.005 \\
    & driver-top3 & classification & 0.005 \\
    & driver-position & regression & 0.0005 \\
    \midrule
    \multirow{1}{*}{\texttt{rel-stack}} 
    & post-post-related & recommendation & 0.0005 \\
    \midrule
    \multirow{1}{*}{\texttt{rel-trial}} 
    & study-outcome & classification & 0.0005 \\
    \bottomrule
    \end{tabular}
\end{table}

\clearpage
\section{Additional Results regarding Our Observations on the \ourdata Datasets}
\label{sec:appendix_res}

\subsection{Additional Results regarding Obs 1. }\label{app:obs1}
\begin{figure*}[ht]
    {\includegraphics[width=0.95\linewidth]{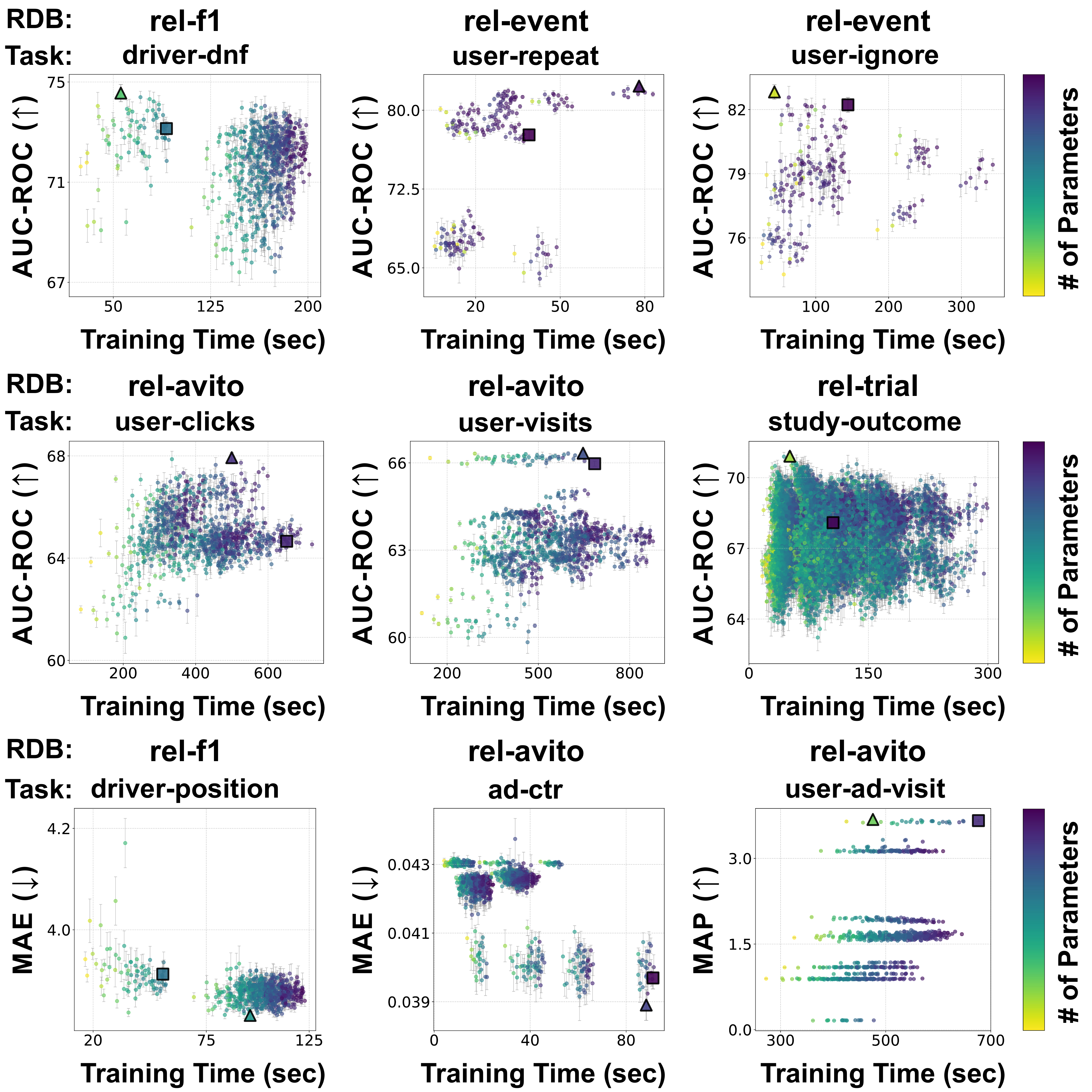}}
    \caption{
        For the nine remaining tasks, we visualize the distribution of performances (Y-axis) across all graph models, along with training time (X-axis) and GNN parameter size (indicated by color).
        Note that there exist graph models yielding substantial improvements in both performance and efficiency compared to those generated by widely-used AR2N modeling~\cite{robinson2024relbench}.
    }
    \label{fig:addperfdist}
\end{figure*}

Figure~\ref{fig:addperfdist} complements Section~\ref{subsec:overallres} by presenting the performance distribution for the remaining 9 tasks.
The results for \driverTopThree, \userAttendance, and \postPostLinked are visualized in Figure~\ref{fig:datasetsummary}~(b) in the main paper.

\clearpage
\subsection{Additional Results regarding Obs 2. }\label{app:obs2}

\begin{figure*}[ht]
    \centering
{\includegraphics[width=0.95\linewidth]{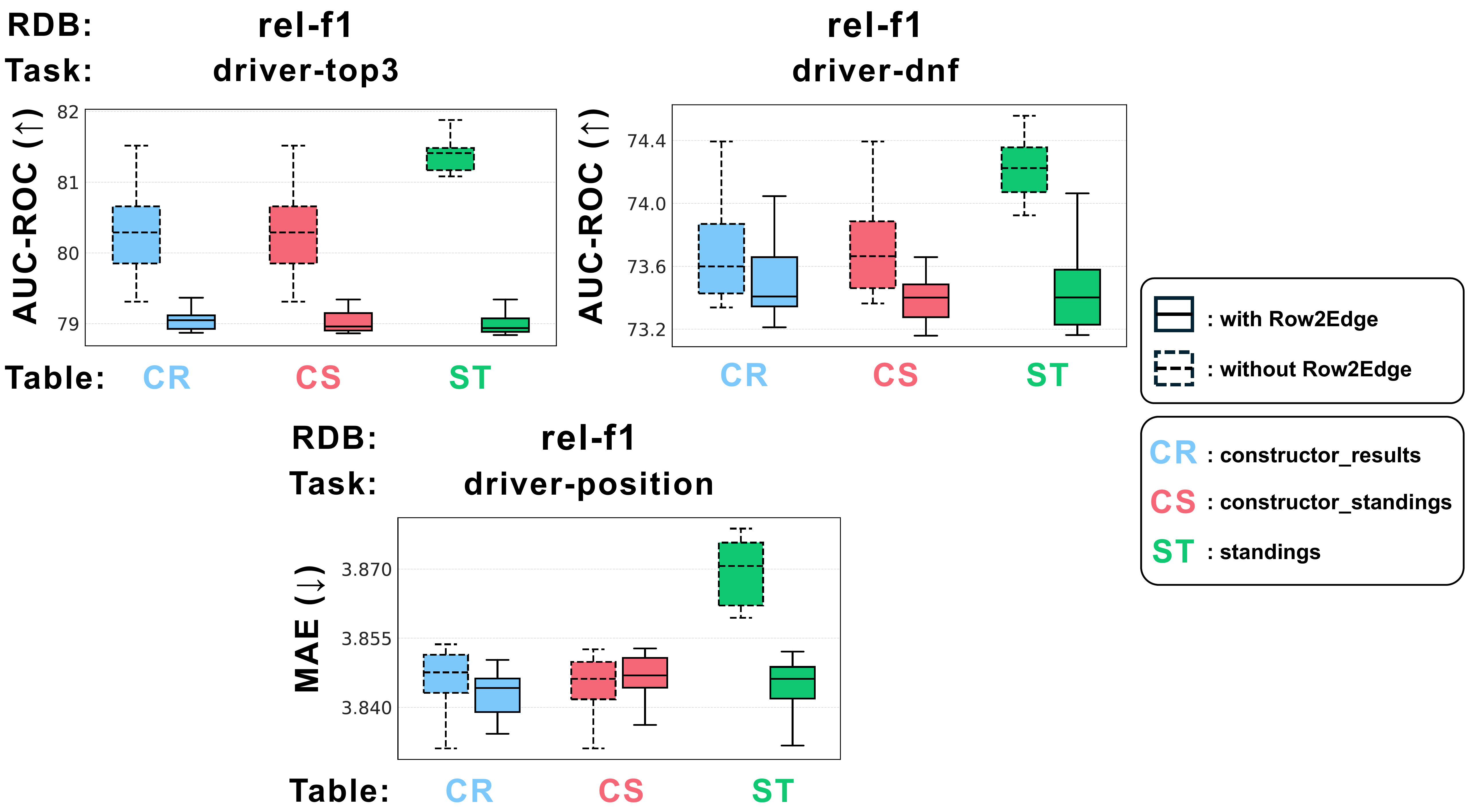}}
    \vspace{0.2cm}
    \caption{
        \textbf{Additional results on \fone regarding Obs 2.}
    }
    \label{fig:row2nef1}
\end{figure*}

\begin{figure*}[ht]
    \vspace{-3mm}
    \centering
{\includegraphics[width=0.95\linewidth]{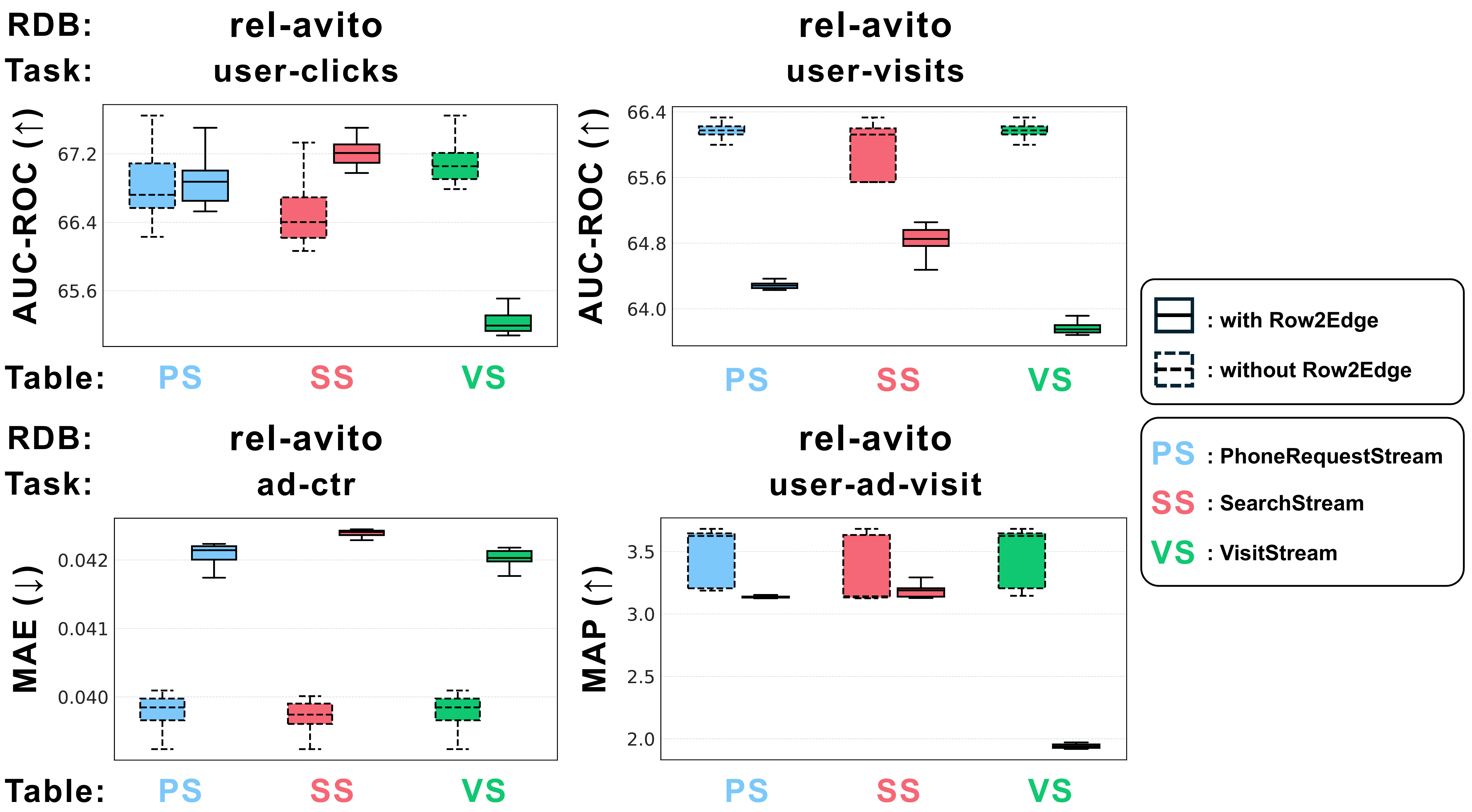}}
    \vspace{0.2cm}
    \caption{
        \textbf{Additional results on \avito regarding Obs 2.} 
    }
    \label{fig:row2neavito}
\end{figure*}

\begin{figure*}[ht]
    \vspace{-3mm}
    \centering
{\includegraphics[width=0.95\linewidth]{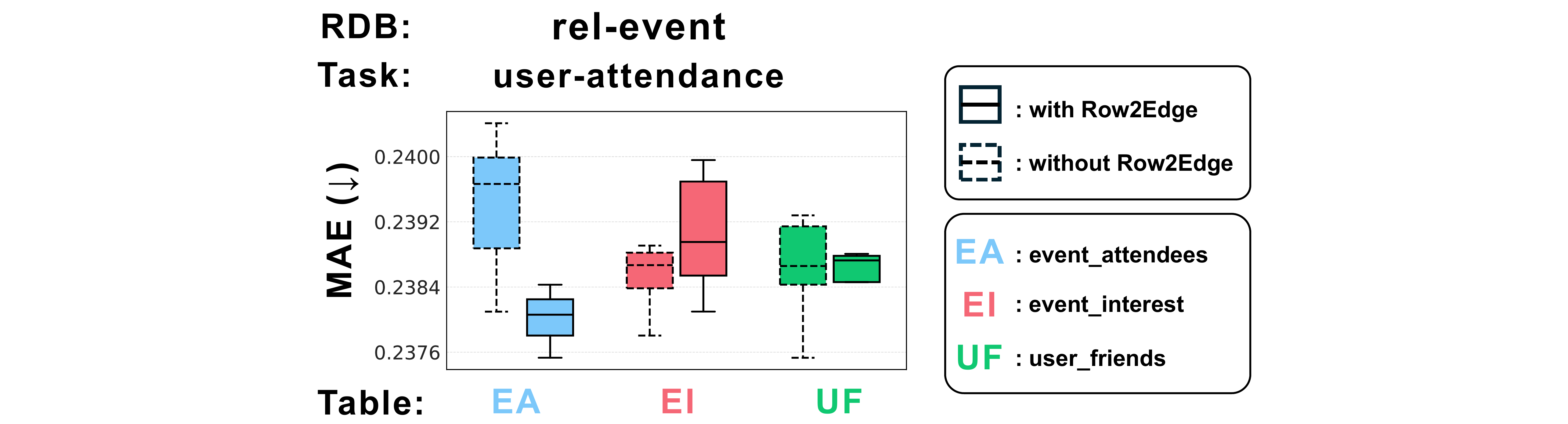}}
    \vspace{0.2cm}
    \caption{
        \textbf{Additional results on \event regarding Obs 2.} 
    }
    \label{fig:row2neevent}
\end{figure*}

\begin{figure*}[ht]
    \vspace{-3mm}
    \centering
{\includegraphics[width=0.95\linewidth]{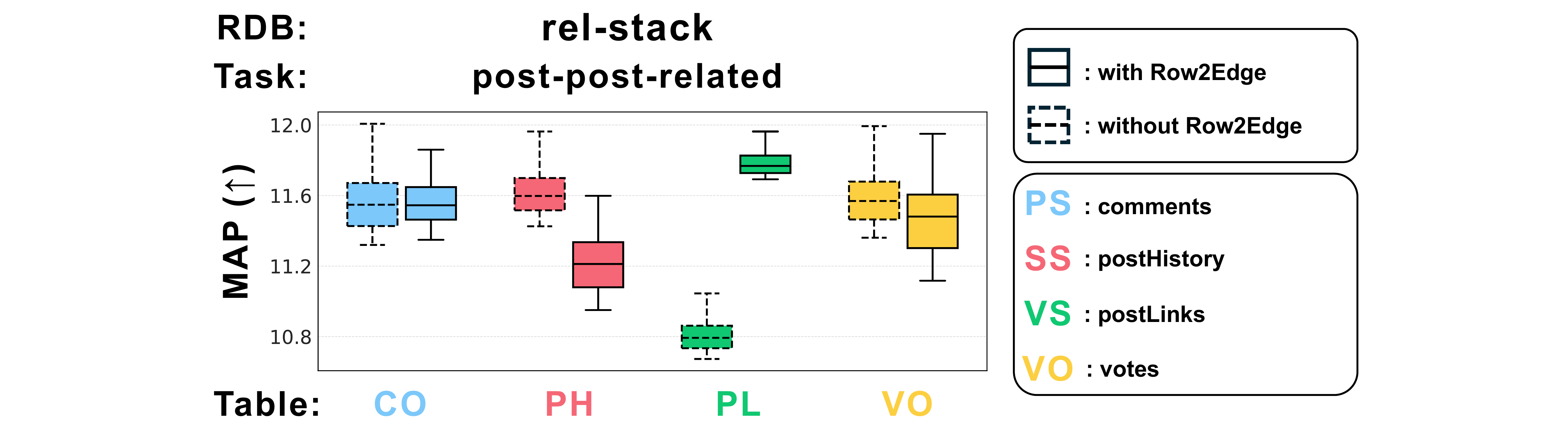}}
    \vspace{0.2cm}
    \caption{
        \textbf{Additional results on \stackex regarding Obs 2.} 
    }
    \label{fig:row2nestack}
\end{figure*}

\begin{figure*}[ht]
    \vspace{-3mm}
    \centering
{\includegraphics[width=0.95\linewidth]{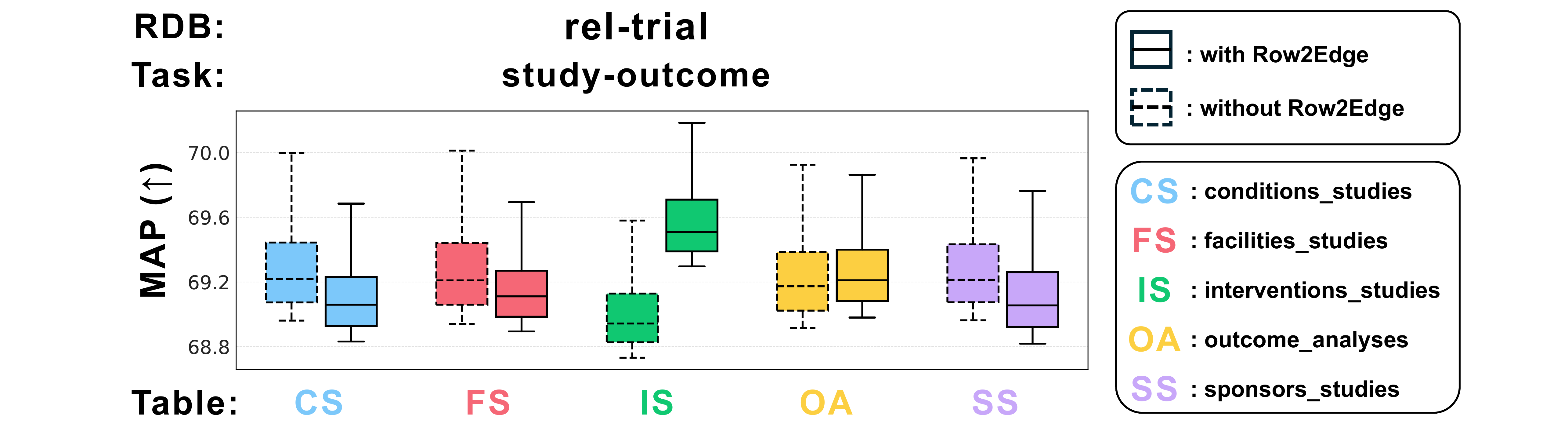}}
    \vspace{0.2cm}
    \caption{
        \textbf{Additional results on \trials regarding Obs 2.} 
    }
    \label{fig:row2netrial}
\end{figure*}

% In this part, we provide additional experimental results supporting Section~\ref{subsec:row2necomparison}, which examines how modeling table rows can lead to markedly different outcomes depending on the specific downstream task. Fig~\ref{fig:row2nef1} - \ref{fig:row2netrial} show the results on the remaining 10 tasks. Note that \userRepeat and \userIgnore are already visualized in Figure~\ref{fig:row2nemain}.
In this section, we provide additional experimental results supporting Section~\ref{subsec:row2necomparison}, which examines how modeling table rows can lead to markedly different outcomes depending on the specific downstream task.
Figures~\ref{fig:row2nef1} - \ref{fig:row2netrial} show the results on the remaining 10 tasks. The results for \userRepeat and \userIgnore are visualized in Figure~\ref{fig:row2nemain} of the main paper.

\clearpage
\subsection{Additional Results regarding Obs 3. }\label{app:obs3}

% In this section, we present additional case study results for Section~\ref{subsec:commonsubstructure}. 
%Fig~\ref{fig:casef1dt} - \ref{fig:casetrialso} provides the case studies on the remaining 11 tasks.
% Note that \userAttendance is already visualized in Figure~\ref{fig:caseeventua}.

\begin{figure*}[ht]
    {\includegraphics[width=0.95\linewidth]{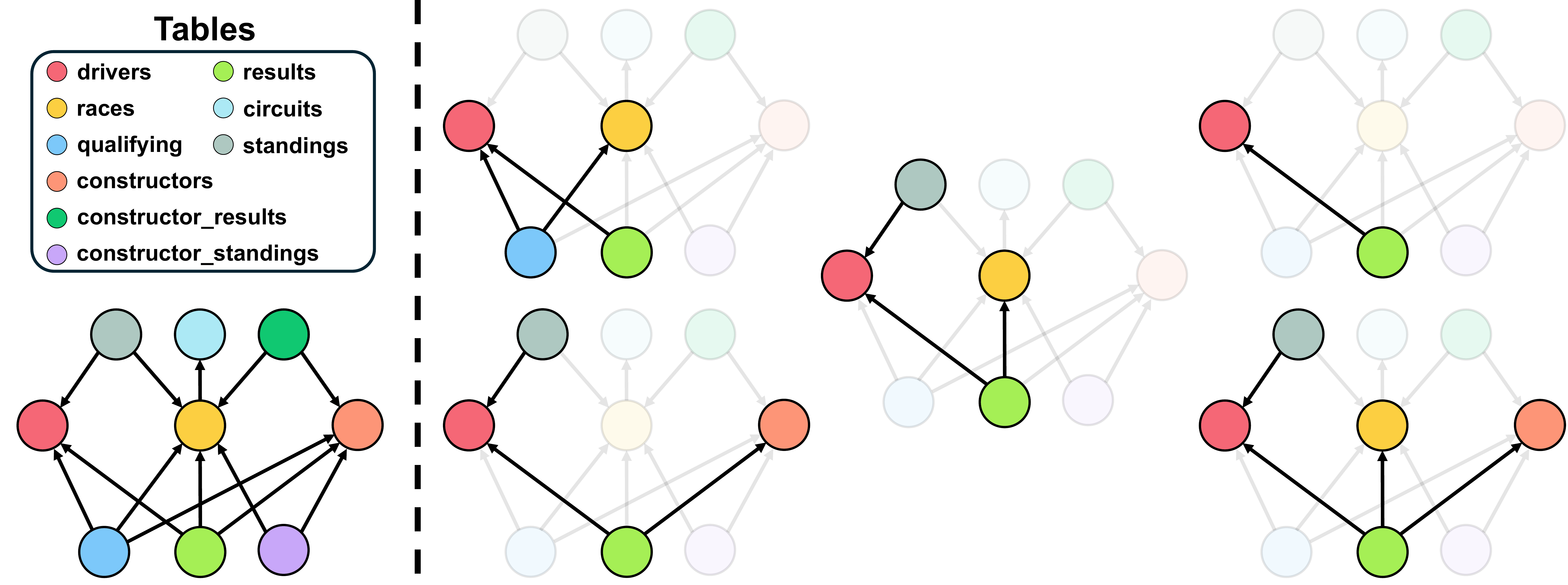}}
    \caption{\textbf{Top performing graph models for \driverTopThree.}
    }
    \label{fig:casef1dt}
\end{figure*}

\begin{figure*}[ht]
    {\includegraphics[width=0.95\linewidth]{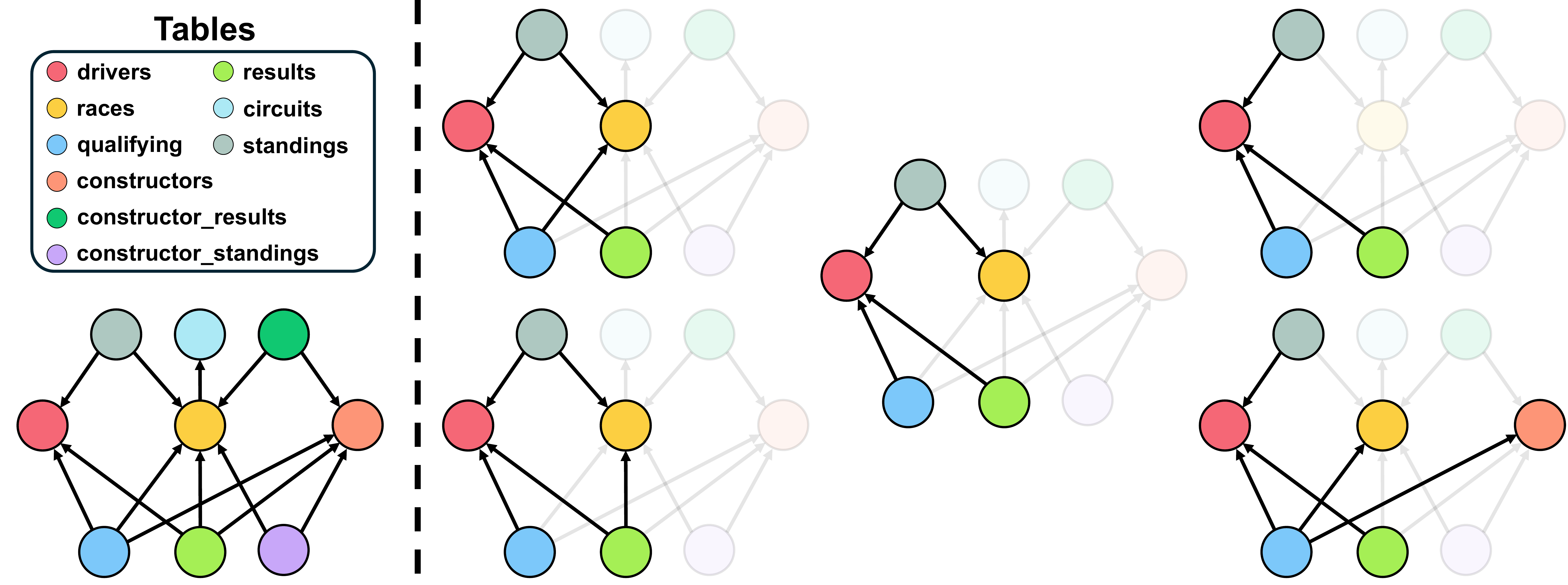}}
    \caption{\textbf{Top performing graph models for \driverDNF.}
    }
    \label{fig:casef1dd}
\end{figure*}

\begin{figure*}[ht]
    {\includegraphics[width=0.95\linewidth]{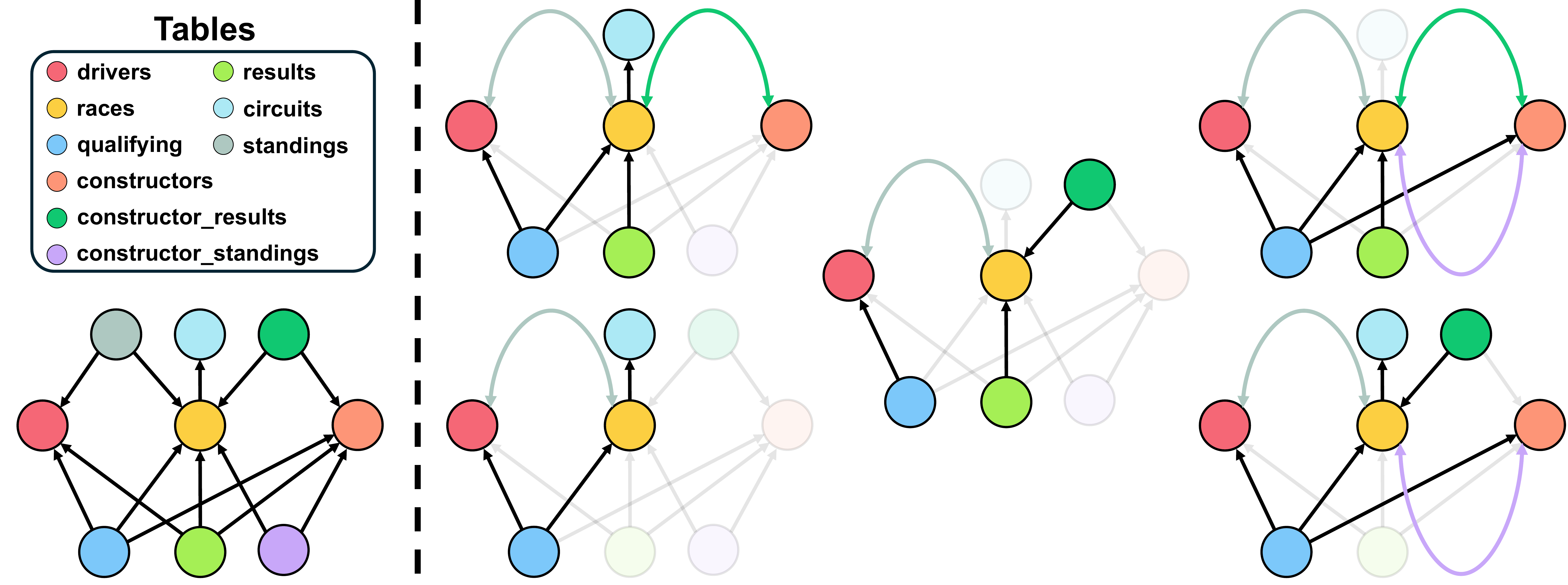}}
    \caption{\textbf{Top performing graph models for \driverPosition.}
    }
    \label{fig:casef1dp}
\end{figure*}

\begin{figure*}[ht]
    {\includegraphics[width=0.95\linewidth]{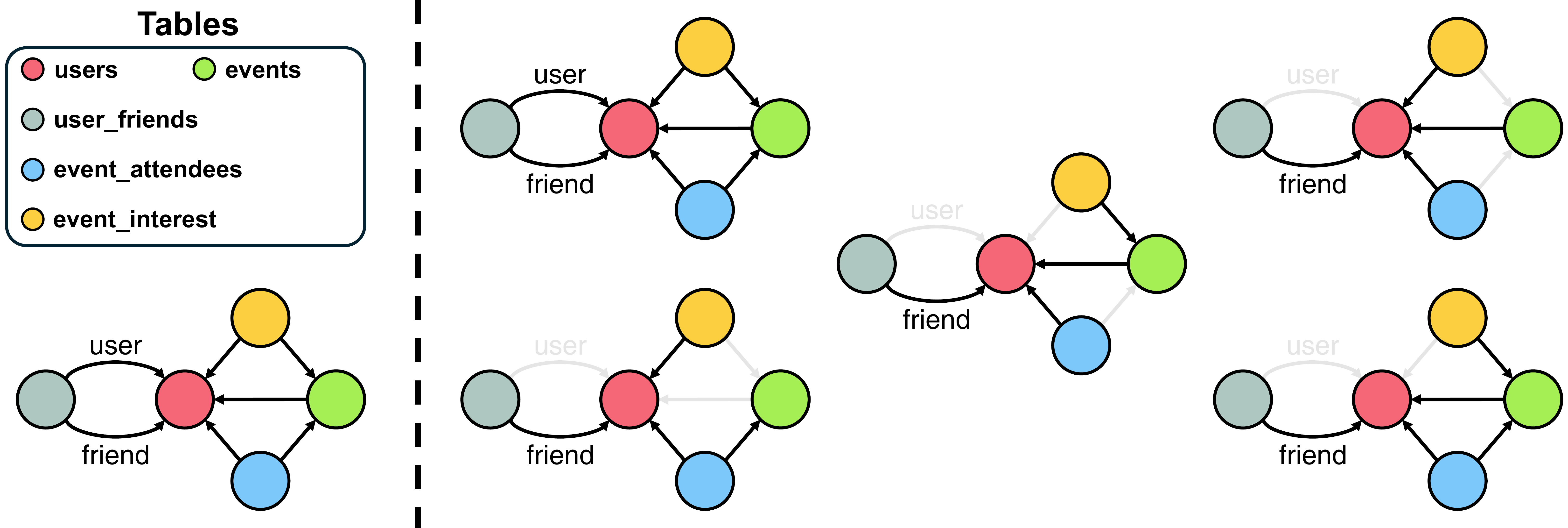}}
    \caption{\textbf{Top performing graph models for  \userIgnore.}
    Note that the \textcolor{case_gray}{users\_friends} table has two FKs (\texttt{user} and \texttt{friend}) both referencing the \textcolor{case_pink}{users} table.
    }
    \label{fig:caseeventui}
\end{figure*}

\begin{figure*}[ht]
    {\includegraphics[width=0.95\linewidth]{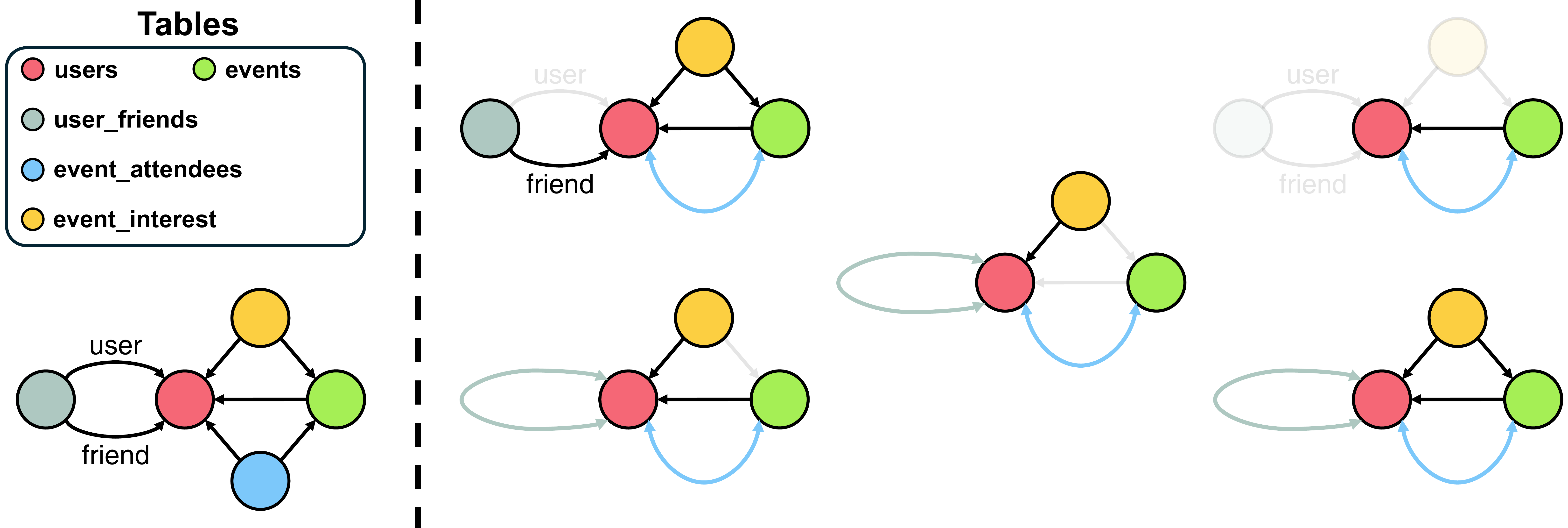}}
    \caption{\textbf{Top performing graph models for \userRepeat.}
    Note that the \textcolor{case_gray}{users\_friends} table has two FKs (\texttt{user} and \texttt{friend}) both referencing the \textcolor{case_pink}{users} table.
    }
    \label{fig:caseeventur}
\end{figure*}

\begin{figure*}[ht]
    {\includegraphics[width=0.95\linewidth]{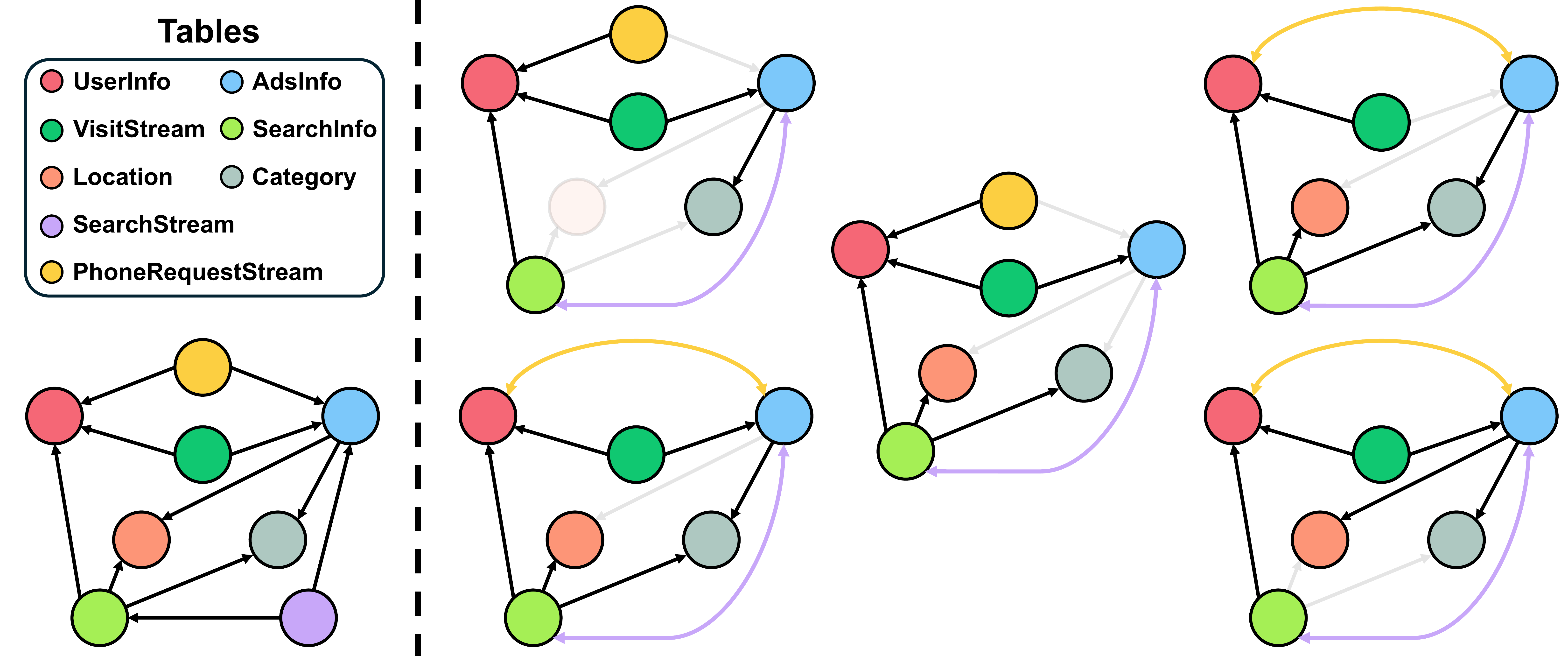}}
    \caption{\textbf{Top performing graph models for \userClick.}}
    \label{fig:caseavitouuc}
\end{figure*}

\begin{figure*}[ht]
    {\includegraphics[width=0.95\linewidth]{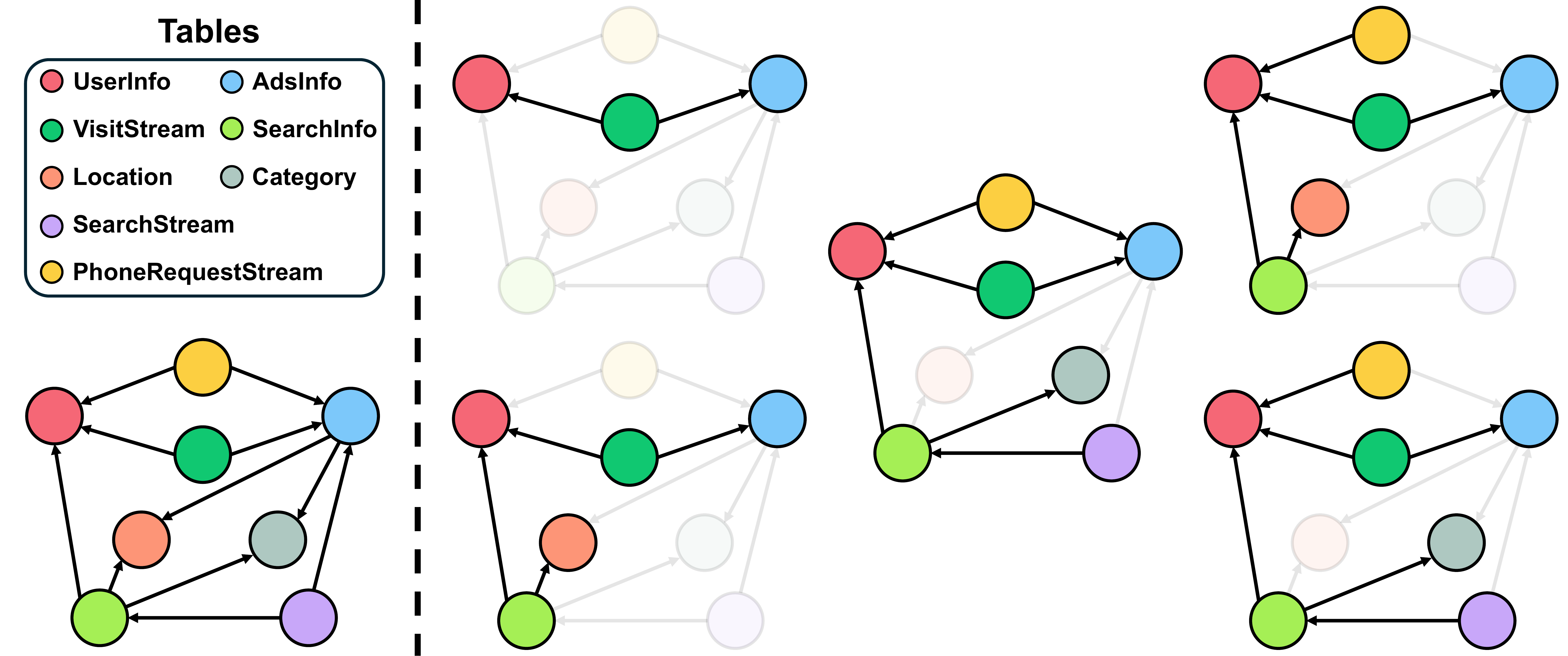}}
    \caption{\textbf{Top performing graph models for \userVisit.}
    }
    \label{fig:caseavitouuv}
\end{figure*}

\begin{figure*}[ht]
    {\includegraphics[width=0.95\linewidth]{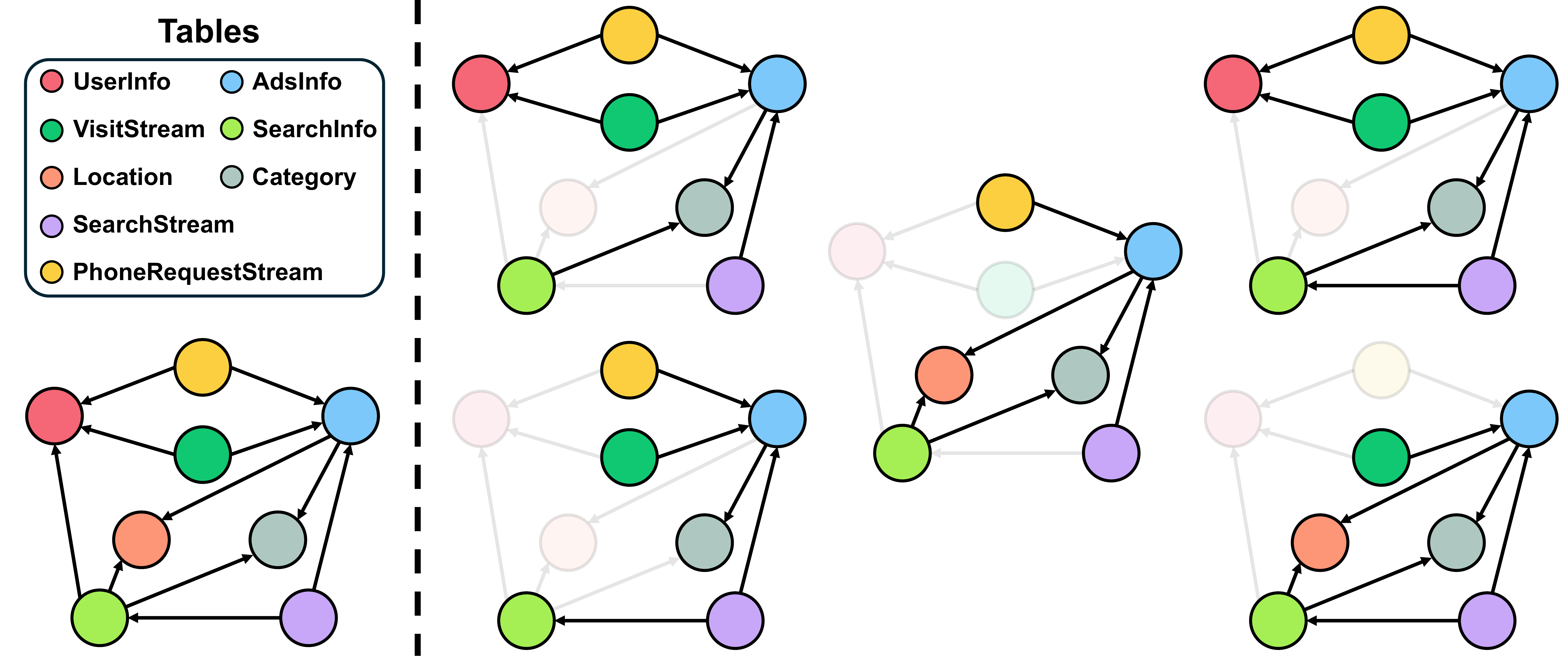}}
    \caption{\textbf{Top performing graph models for \adsCTR.}
    }
    \label{fig:caseavitoac}
\end{figure*}

\begin{figure*}[t]
    {\includegraphics[width=0.95\linewidth]{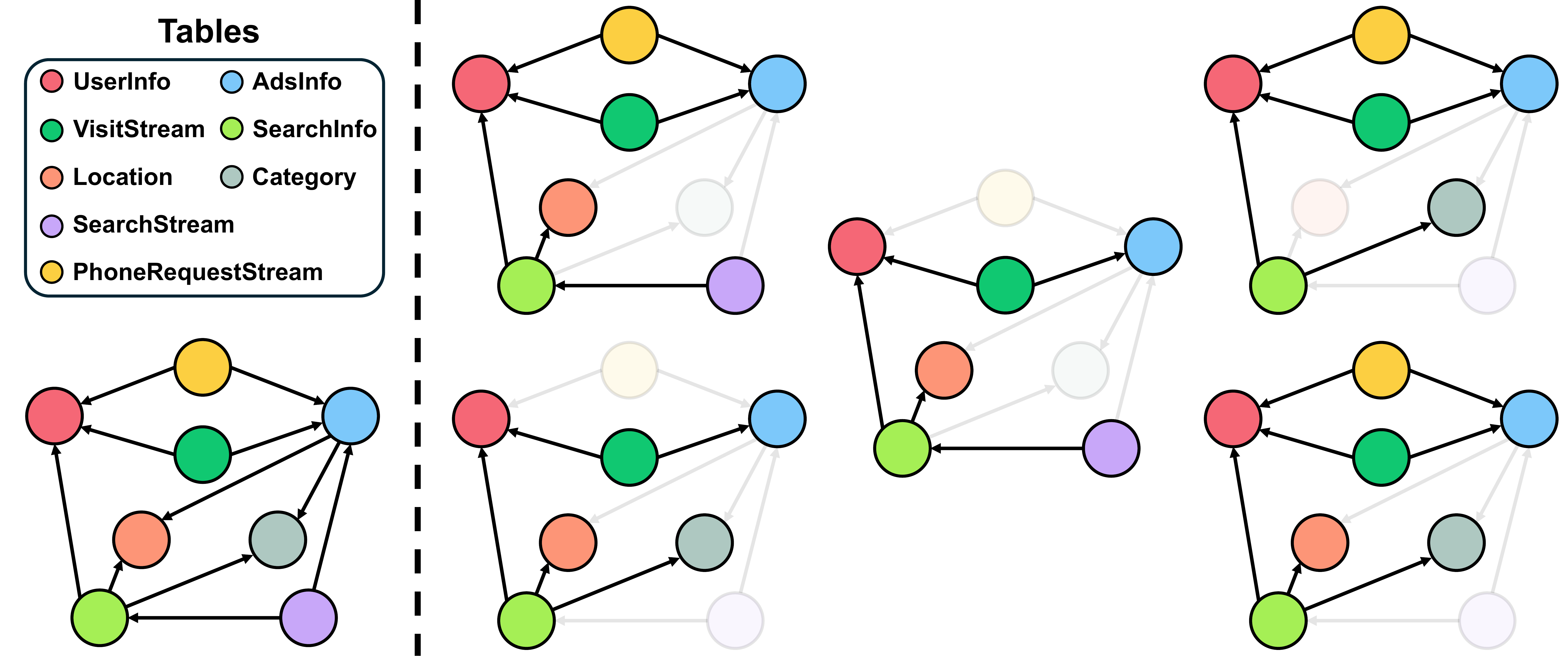}}
    \caption{\textbf{Top performing graph models for \userAdVisit.}
    }
    \label{fig:caseavitouav}
\end{figure*}

\begin{figure*}[ht]
    {\includegraphics[width=0.95\linewidth]{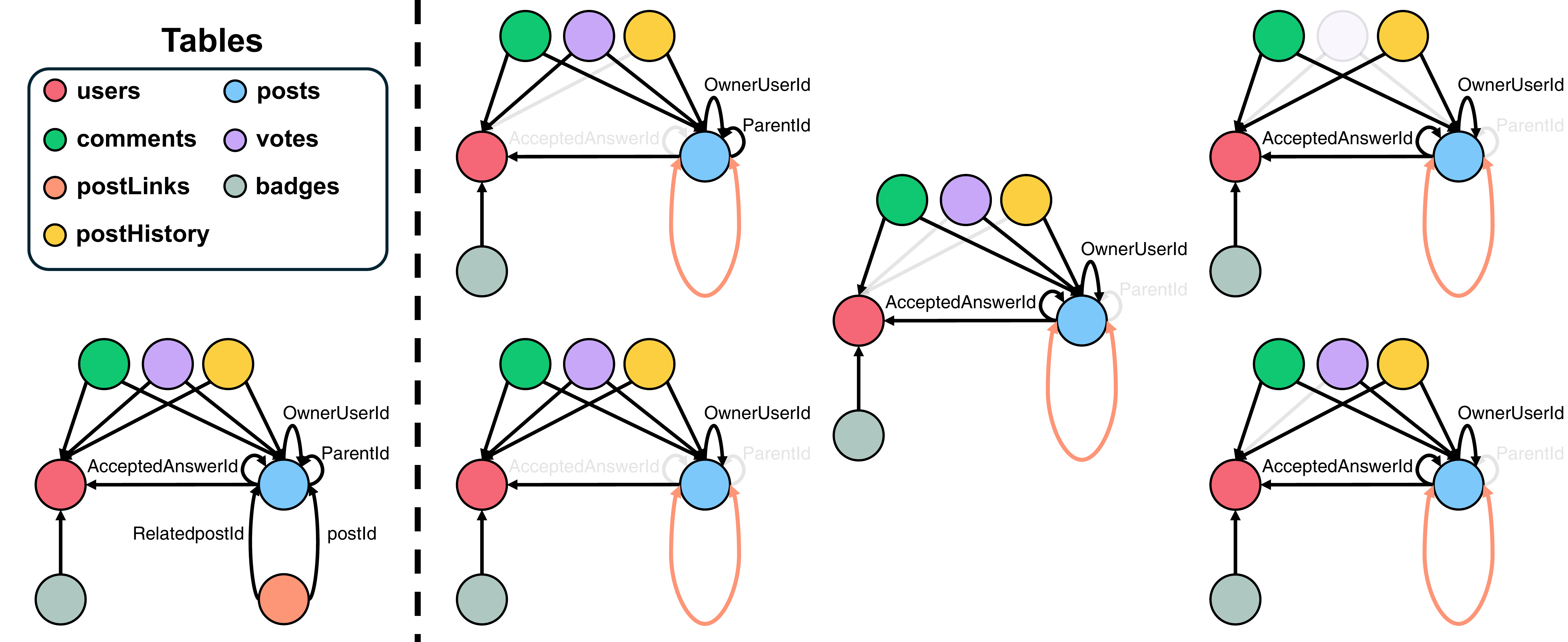}}
    \caption{\textbf{Top performing graph models for \postPostLinked.}
    Note that the \textcolor{case_orange}{postLinks} table has two FKs (\texttt{RelatedpostId} and \texttt{postId}) both referencing the \textcolor{case_blue}{posts} table, and the \textcolor{case_blue}{posts} table has three FKs (\texttt{AcceptedAnswerId}, \texttt{OwnerUserId}, and \texttt{ParentId}) all referencing the table itself.
    }
    \label{fig:casestackppl}
\end{figure*}

\begin{figure*}[ht]
    {\includegraphics[width=0.95\linewidth]{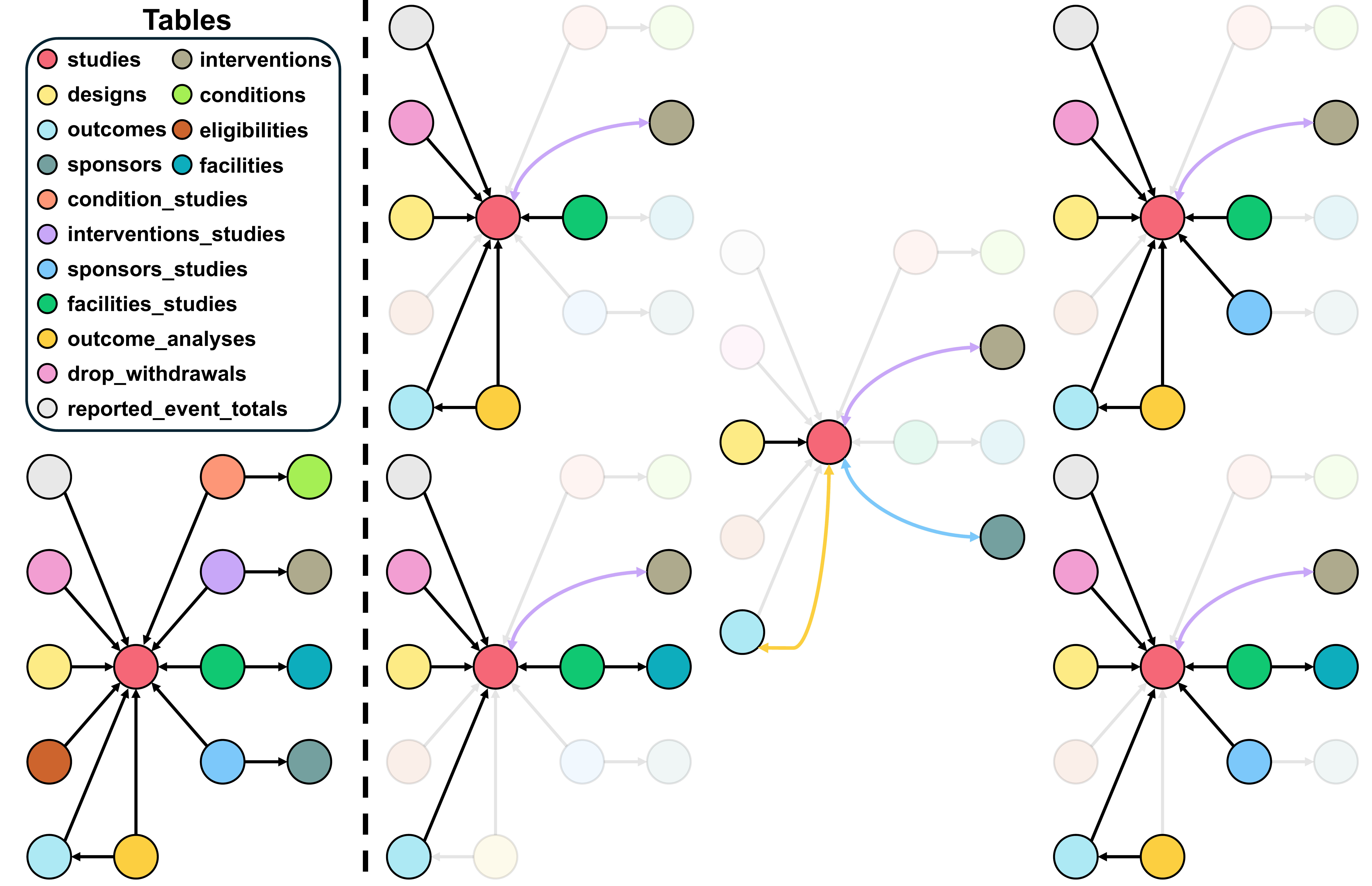}}
    \caption{\textbf{Top performing graph models for \studyOutcome.}
    }
    \label{fig:casetrialso}
\end{figure*}

In this section, we present additional results to supplement Section~\ref{subsec:commonsubstructure}. 
Figures~\ref{fig:casef1dt} - \ref{fig:casetrialso} provide the case study results on the remaining 11 tasks. 
The case study result for \userAttendance is provided in Figure~\ref{fig:caseeventua} in the main paper.

\clearpage
\subsection{Additional Results regarding Obs 5. }\label{app:obs5}

In this section, we provide extra details and results to supplement Section~\ref{subsec:modelgeneralization}.
Given our computational constraints, we examine only the correlations for the top- and bottom-5\% graph models—ranked by GraphSAGE (sum aggregation)—across three new predictors (i.e., graph neural networks): GraphSAGE (mean aggregation), GIN, and GPS.

\begin{figure*}[ht]
    \centering
    {\includegraphics[width=0.95\linewidth]{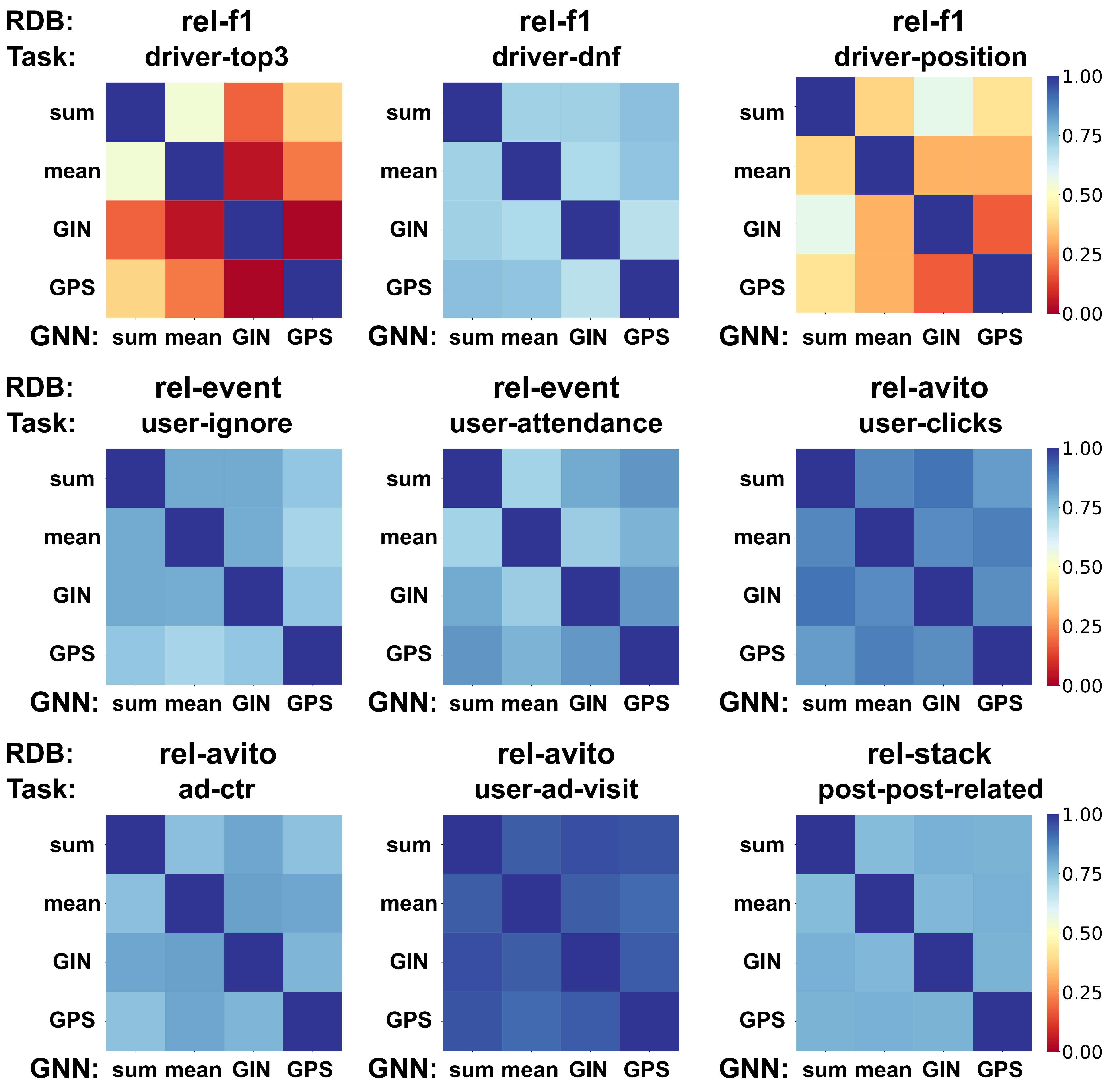}}
    \caption{
        \textbf{Additional results regarding Obs 5.} Spearman correlations between different predictive GNNs—GraphSAGE with sum and mean aggregation (denoted as sum and mean), GIN, and GPS—are generally high (above 0.7) across the remaining tasks.
    }
    \label{fig:analysismodel}
\end{figure*}

As shown in Figure~\ref{fig:analysismodel}, while
cross-GNN correlations are relatively low for the two tasks
on the \fone dataset (\driverTopThree and \driverPosition) due to high variances in the performances, for the other tasks, cross-GNN correlations are high.
The results for \userRepeat, \userVisit, and \userAdVisit are presented in Figure~\ref{fig:modelgeneralization} in the main paper.

\section{Benchmark Detail of the \ourdata}
\label{sec:appendix_bench}
\subsection{Implementation Details of Baselines} \label{app:baselinedetail}
In this section, we provide implementation details for the action-based search algorithms: Evolutionary Algorithm (EA), Bayesian Optimization (BO), and Reinforcement Learning (RL). %Their hyperparameters were briefly tuned to achieve reasonably comparable performance to other baselines.

\textbf{Evolutionary Algorithm. } The Evolutionary Algorithm (EA) baseline employs a regularized evolutionary search strategy. The algorithm initializes a population by randomly sampling graph configurations and iteratively evolves this population through mutation and selection processes. At each iteration, a subset of individuals is selected through tournament selection, and mutations are applied using defined micro-actions to generate offspring. The oldest individual in the population is replaced by the best-performing offspring, maintaining diversity and continuous exploration.

Its key hyperparameter settings are summarized below:

\begin{itemize}
\item \textbf{Population Size}: $\min(10, \text{budget})$
\item \textbf{Tournament Size}: $\min(10, \text{budget})$
\item \textbf{Max Iterations}: 1000
\end{itemize}

\textbf{Bayesian Optimization. } The Bayesian Optimization (BO) baseline employs an iterative search leveraging a surrogate Multi-Layer Perceptron (MLP) based on the BANANAS~\cite{white2021bananas} implementation. The input embeddings represent graph structures converted into fixed-sized arrays based on selected edges.

The loss function follows the original BANANAS formulation:

\begin{equation}
L = \text{mean}\left| \frac{y_{\text{pred}} - y_{\text{lb}}}{y_{\text{true}} - y_{\text{lb}}} - 1 \right|
\end{equation}

Initial embeddings are obtained through random sampling, and subsequent selections are guided by Expected Improvement, computed via Monte Carlo sampling with 50 samples. This iterative process continues until a defined budget limit or a maximum of 100 iterations is reached.

Its key hyperparameter settings are summarized below:

\begin{itemize}
\item \textbf{Surrogate Model}: MLP (2 hidden layers, each with 32 units)
\item \textbf{Dropout Rate}: 0.1
\item \textbf{Optimizer}: Adam
\item \textbf{Learning Rate}: 0.001
\item \textbf{Epochs per Iteration}: 50
\item \textbf{Batch Size}: 32
\item \textbf{Initial Sampling Size}: $\min(10, \text{budget})$
\item \textbf{EI MC Samples}: 50
\item \textbf{Max Iterations}: 100
\end{itemize}

\textbf{Reinforcement Learning. }
The Reinforcement Learning (RL) baseline utilizes a policy gradient approach with an RNN-based controller for the search process.
The controller is implemented as a one-layer LSTM~\cite{hochreiter1997long} with 32 hidden units. Input state embeddings are derived from the current graph structures encoded as fixed-size vectors.

Each training episode consists of up to five steps, during which the controller selects actions based on the policy generated from the RNN outputs.
Rewards are computed based on performance improvements between sequential states, using a discount factor of 0.99.
Training runs for up to 50 episodes or until a predefined evaluation budget is reached.
Its key hyperparameter settings are summarized below:

\begin{itemize}
\item \textbf{Controller Model}: LSTM (1 layer, 32 hidden units)
\item \textbf{Optimizer}: Adam
\item \textbf{Learning Rate}: 0.005
\item \textbf{Max Steps per Episode}: 5
\item \textbf{Discount Factor ($\gamma$)}: 0.99
\item \textbf{Episodes}: 50
\end{itemize}

\section{Additional Results regarding Benchmark on the \ourdata Datasets}
\label{sec:appendix_benchres}
\subsection{Performance Results on the \ourdata Datasets} \label{app:benchresults}
\begin{figure*}[!ht]
    \centering
    {\includegraphics[width=0.95\linewidth]{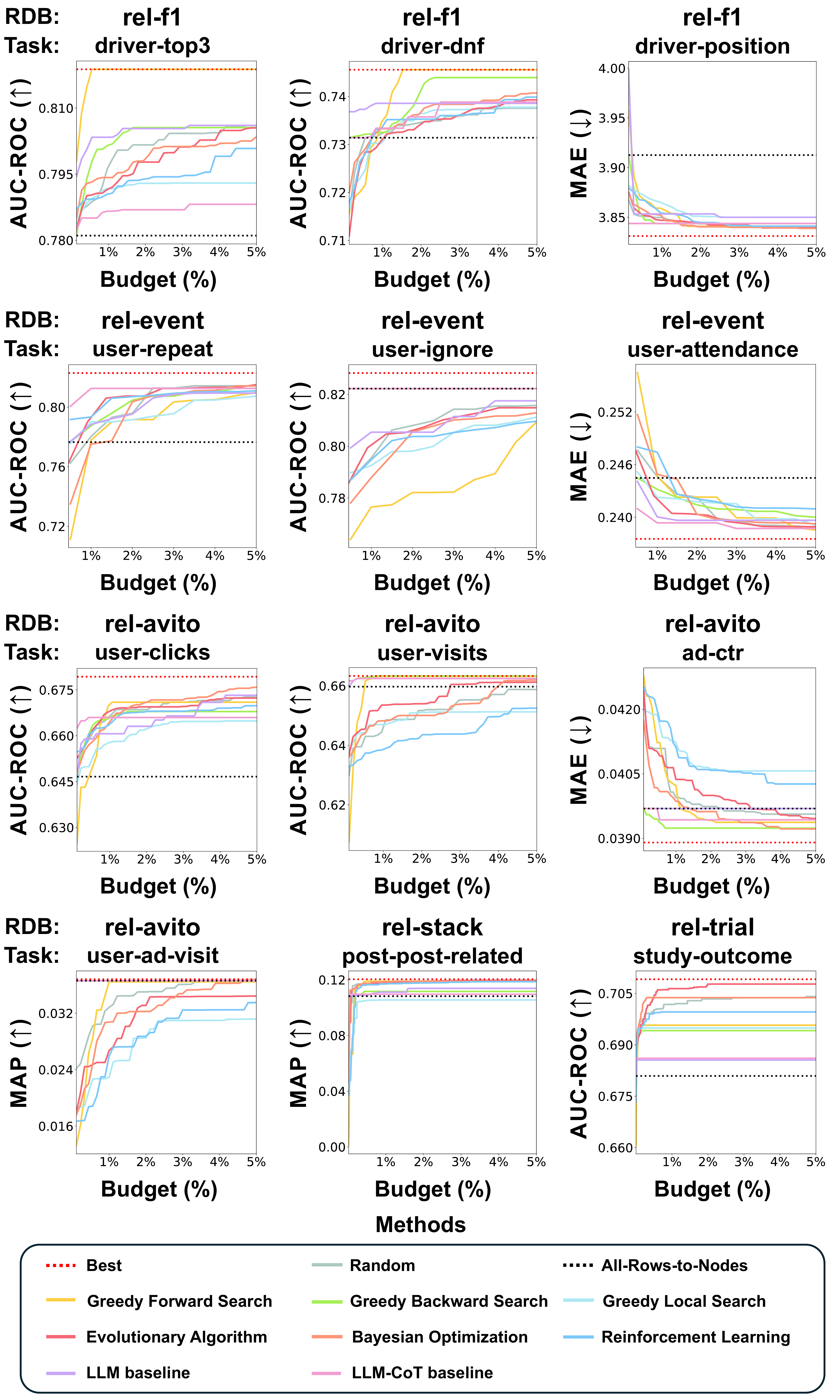}}
    \caption{
        Performance details of ten RDB-to-graph modeling methods on each of the 12 predictive tasks.
        The performances are computed under varying budget levels, corresponding to the number of graph models evaluated.
    }
    \label{fig:addbenchresult}
\end{figure*}

In this section, we provide additional performance details of ten RDB-to-graph modeling methods, supplementing the summary presented in Table~\ref{tab:benchfone} and Figure~\ref{fig:benchmark_rank} in the main paper.
Details are presented in Tables~\ref{tab:benchevent} - \ref{tab:benchstacktrial} and Figure~\ref{fig:addbenchresult}.

\begin{table}[ht]
  \centering
  \caption{Performance of ten RDB-to-graph modeling methods on the \event dataset under varying budget levels.}
  \label{tab:benchevent}
  \setlength{\tabcolsep}{3pt}
  \vspace{1mm}
  \scalebox{0.73}{
  \begin{tabular}{l|ccccc|ccccc|ccccc}
    \toprule
    \textbf{Task Name} &
    \multicolumn{5}{|c|}{\textbf{\userIgnore\ (AUC-ROC (\%) $\uparrow$)}} &
    \multicolumn{5}{c|}{\textbf{\userRepeat\ (AUC-ROC (\%) $\uparrow$)}} &
    \multicolumn{5}{c}{\textbf{\userAttendance\ (MSE $\downarrow$)}} \\
    \midrule
    \mr{2}{\textbf{Methods}} & \multicolumn{5}{c|}{\textbf{Budget (\%)}} &
    \multicolumn{5}{c|}{\textbf{Budget (\%)}} &
    \multicolumn{5}{c}{\textbf{Budget (\%)}} \\
    & \textbf{1\%} & \textbf{2\%} & \textbf{3\%} & \textbf{4\%} & \textbf{5\%}
    & \textbf{1\%} & \textbf{2\%} & \textbf{3\%} & \textbf{4\%} & \textbf{5\%}
    & \textbf{1\%} & \textbf{2\%} & \textbf{3\%} & \textbf{4\%} & \textbf{5\%} \\
    \midrule\midrule
    Best     & 82.823 & 82.823 & 82.823 & 82.823 & 82.823
             & 82.291 & 82.291 & 82.291 & 82.291 & 82.291
             & 0.2375 & 0.2375 & 0.2375 & 0.2375 & 0.2375 \\
    \midrule
    Random   & 79.543 & 80.797 & 81.427 & 81.548 & 81.572
             & 78.034 & 79.588 & 81.280 & 81.430 & 81.430
             & 0.2446 & 0.2423 & 0.2396 & 0.2391 & 0.2389 \\
    AR2N     & 82.222 & 82.222 & 82.222 & 82.222 & 82.222
             & 77.651 & 77.651 & 77.651 & 77.651 & 77.651
             & 0.2445 & 0.2445 & 0.2445 & 0.2445 & 0.2445 \\
    GF       & 77.655 & 78.213 & 78.233 & 78.945 & 80.931
             & 77.789 & 79.163 & 80.348 & 80.508 & 80.981
             & 0.2447 & 0.2423 & 0.2399 & 0.2399 & 0.2385 \\
    GB       & 82.222 & 82.222 & 82.222 & 82.222 & 82.222
             & 78.748 & 80.469 & 80.744 & 81.018 & 81.258
             & 0.2432 & 0.2414 & 0.2409 & 0.2407 & 0.2400 \\
    GL       & 79.256 & 79.809 & 80.534 & 80.809 & 81.120
             & 79.012 & 79.158 & 79.558 & 80.481 & 80.739
             & 0.2423 & 0.2418 & 0.2415 & 0.2398 & 0.2391 \\
    EA       & 79.939 & 80.523 & 81.011 & 81.299 & 81.482
             & 78.989 & 80.739 & 81.249 & 81.313 & 81.333
             & 0.2425 & 0.2404 & 0.2395 & 0.2390 & 0.2389 \\
    BO       & 78.787 & 80.550 & 81.015 & 81.162 & 81.288
             & 77.516 & 80.360 & 80.831 & 81.120 & 81.424
             & 0.2449 & 0.2403 & 0.2396 & 0.2394 & 0.2392 \\
    RL       & 79.474 & 80.376 & 80.527 & 80.719 & 80.963
             & 79.341 & 80.703 & 80.930 & 81.040 & 81.094
             & 0.2474 & 0.2421 & 0.2412 & 0.2410 & 0.2409 \\
    LLM      & 80.541 & 80.541 & 81.146 & 81.751 & 81.751
             & 78.841 & 79.537 & 80.857 & 80.941 & 80.941
             & 0.2400 & 0.2397 & 0.2397 & 0.2397 & 0.2397 \\
    LLM-CoT  & 82.222 & 82.222 & 82.222 & 82.222 & 82.222
             & 81.259 & 81.259 & 81.259 & 81.259 & 81.259
             & 0.2394 & 0.2394 & 0.2387 & 0.2387 & 0.2387 \\
    \bottomrule
  \end{tabular}
  } % end resizebox
  \vspace{2mm}
\end{table}

\begin{table}[ht]
  \centering
  \caption{Performance of ten RDB-to-graph modeling methods on the \avito dataset under varying budget levels.}
  \label{tab:benchavito}
  \setlength{\tabcolsep}{3pt}
  \vspace{1mm}

  \scalebox{0.73}{
  \begin{tabular}{l|ccccc|ccccc}
    \toprule
    \textbf{Task Name} &
    \multicolumn{5}{|c|}{\textbf{\userClick\ (AUC-ROC (\%) $\uparrow$)}} &
    \multicolumn{5}{c}{\textbf{\userVisit\ (AUC-ROC (\%) $\uparrow$)}} \\
    \midrule
    \mr{2}{\textbf{Methods}} &
    \multicolumn{5}{c|}{\textbf{Budget (\%)}} &
    \multicolumn{5}{c}{\textbf{Budget (\%)}} \\
    & \textbf{1\%} & \textbf{2\%} & \textbf{3\%} & \textbf{4\%} & \textbf{5\%}
    & \textbf{1\%} & \textbf{2\%} & \textbf{3\%} & \textbf{4\%} & \textbf{5\%} \\
    \midrule\midrule
    Best     & 67.931 & 67.931 & 67.931 & 67.931 & 67.931
             & 66.332 & 66.332 & 66.332 & 66.332 & 66.332 \\
    \midrule
    Random   & 66.594 & 66.856 & 67.098 & 67.151 & 67.302
             & 64.444 & 65.169 & 65.521 & 65.708 & 65.877 \\
    AR2N     & 64.660 & 64.660 & 64.660 & 64.660 & 64.660
             & 65.971 & 65.971 & 65.971 & 65.971 & 65.971 \\
    GF       & 66.978 & 67.096 & 67.096 & 67.096 & 67.096
             & 66.318 & 66.318 & 66.318 & 66.318 & 66.318 \\
    GB       & 66.561 & 66.791 & 66.791 & 66.791 & 66.791
             & 66.332 & 66.332 & 66.332 & 66.332 & 66.332 \\
    GL       & 65.770 & 66.183 & 66.468 & 66.468 & 66.483
             & 64.700 & 65.090 & 65.118 & 65.124 & 65.124 \\
    EA       & 66.778 & 66.944 & 66.958 & 67.204 & 67.232
             & 65.352 & 65.397 & 66.047 & 66.116 & 66.122 \\
    BO       & 66.388 & 67.069 & 67.170 & 67.442 & 67.584
             & 64.823 & 65.004 & 65.380 & 65.924 & 66.191 \\
    RL       & 66.341 & 66.771 & 66.801 & 66.902 & 66.983
             & 63.840 & 64.228 & 64.379 & 64.963 & 65.249 \\
    LLM      & 66.066 & 66.308 & 66.648 & 67.092 & 67.323
             & 66.246 & 66.246 & 66.246 & 66.246 & 66.246 \\
    LLM-CoT  & 66.597 & 66.597 & 66.597 & 66.597 & 66.597
             & 66.246 & 66.246 & 66.246 & 66.246 & 66.246 \\
    \bottomrule
  \end{tabular}
  }

  \vspace{3mm}

  \scalebox{0.73}{
  \begin{tabular}{l|ccccc|ccccc}
    \toprule
    \textbf{Task Name} &
    \multicolumn{5}{|c|}{\textbf{\adsCTR\ (MSE $\downarrow$)}} &
    \multicolumn{5}{c}{\textbf{\userAdVisit\ (MAP (\%) $\uparrow$)}} \\
    \midrule
    \mr{2}{\textbf{Methods}} &
    \multicolumn{5}{c|}{\textbf{Budget (\%)}} &
    \multicolumn{5}{c}{\textbf{Budget (\%)}} \\
    & \textbf{1\%} & \textbf{2\%} & \textbf{3\%} & \textbf{4\%} & \textbf{5\%}
    & \textbf{1\%} & \textbf{2\%} & \textbf{3\%} & \textbf{4\%} & \textbf{5\%} \\
    \midrule\midrule
    Best     & 0.0389 & 0.0389 & 0.0389 & 0.0389 & 0.0389
             & 3.6816 & 3.6816 & 3.6816 & 3.6816 & 3.6816 \\
    \midrule
    Random   & 0.0399 & 0.0397 & 0.0396 & 0.0396 & 0.0396
             & 3.2824 & 3.5017 & 3.6084 & 3.6468 & 3.6487 \\
    AR2N     & 0.0397 & 0.0397 & 0.0397 & 0.0397 & 0.0397
             & 3.6610 & 3.6610 & 3.6610 & 3.6610 & 3.6610 \\
    GF       & 0.0402 & 0.0394 & 0.0394 & 0.0394 & 0.0394
             & 3.6453 & 3.6453 & 3.6453 & 3.6453 & 3.6453 \\
    GB       & 0.0392 & 0.0392 & 0.0392 & 0.0392 & 0.0392
             & 3.6610 & 3.6610 & 3.6610 & 3.6610 & 3.6610 \\
    GL       & 0.0412 & 0.0406 & 0.0406 & 0.0406 & 0.0406
             & 2.2905 & 2.8499 & 3.0970 & 3.0989 & 3.1171 \\
    EA       & 0.0404 & 0.0400 & 0.0398 & 0.0395 & 0.0395
             & 2.6709 & 3.3097 & 3.4353 & 3.4391 & 3.4432 \\
    BO       & 0.0399 & 0.0396 & 0.0394 & 0.0392 & 0.0392
             & 3.0748 & 3.2248 & 3.4774 & 3.6241 & 3.6759 \\
    RL       & 0.0411 & 0.0406 & 0.0405 & 0.0403 & 0.0403
             & 2.5915 & 2.8723 & 3.2453 & 3.2475 & 3.3514 \\
    LLM      & 0.0397 & 0.0397 & 0.0397 & 0.0397 & 0.0397
             & 3.6610 & 3.6610 & 3.6610 & 3.6610 & 3.6610 \\
    LLM-CoT  & 0.0394 & 0.0394 & 0.0394 & 0.0394 & 0.0394
             & 3.6610 & 3.6634 & 3.6655 & 3.6655 & 3.6679 \\
    \bottomrule
  \end{tabular}
  }
  \vspace{2mm}
\end{table}

\begin{table}[ht]
  \centering
  \caption{Performance of ten RDB-to-graph modeling methods on the \stackex dataset (left) and the \trials dataset (right) under varying budget levels.}
  \label{tab:benchstacktrial}
  \setlength{\tabcolsep}{3pt}
  \vspace{1mm}
  \scalebox{0.73}{
  \begin{tabular}{l|ccccc|ccccc}
    \toprule
    \textbf{Task Name} &
    \multicolumn{5}{|c|}{\textbf{\postPostLinked\ (MAP (\%) $\uparrow$)}} &
    \multicolumn{5}{c}{\textbf{\studyOutcome\ (AUC-ROC (\%) $\uparrow$)}} \\
    \midrule
    \mr{2}{\textbf{Methods}} &
    \multicolumn{5}{c|}{\textbf{Budget (\%)}} &
    \multicolumn{5}{c}{\textbf{Budget (\%)}} \\
    & \textbf{1\%} & \textbf{2\%} & \textbf{3\%} & \textbf{4\%} & \textbf{5\%}
    & \textbf{1\%} & \textbf{2\%} & \textbf{3\%} & \textbf{4\%} & \textbf{5\%} \\
    \midrule\midrule
    Best     & 12.040 & 12.040 & 12.040 & 12.040 & 12.040
             & 70.913 & 70.913 & 70.913 & 70.913 & 70.913 \\
    \midrule
    Random   & 11.751 & 11.803 & 11.835 & 11.867 & 11.900
             & 70.178 & 70.338 & 70.372 & 70.375 & 70.407 \\
    AR2N     & 10.823 & 10.823 & 10.823 & 10.823 & 10.823
             & 68.091 & 68.091 & 68.091 & 68.091 & 68.091 \\
    GF       & 11.903 & 11.903 & 11.903 & 11.903 & 11.903
             & 69.572 & 69.572 & 69.572 & 69.572 & 69.572 \\
    GB       & 11.165 & 11.165 & 11.165 & 11.165 & 11.165
             & 69.413 & 69.413 & 69.413 & 69.413 & 69.413 \\
    GL       & 10.555 & 10.555 & 10.555 & 10.555 & 10.555
             & 69.490 & 69.490 & 69.490 & 69.490 & 69.490 \\
    EA       & 11.786 & 11.897 & 11.946 & 11.954 & 11.954
             & 70.639 & 70.771 & 70.772 & 70.772 & 70.772 \\
    BO       & 11.860 & 11.888 & 11.888 & 11.888 & 11.888
             & 70.376 & 70.376 & 70.376 & 70.376 & 70.376 \\
    RL       & 11.706 & 11.776 & 11.826 & 11.863 & 11.863
             & 69.959 & 69.959 & 69.959 & 69.959 & 69.959 \\
    LLM      & 11.111 & 11.386 & 11.386 & 11.386 & 11.386
             & 68.556 & 68.556 & 68.556 & 68.556 & 68.556 \\
    LLM-CoT  & 10.964 & 10.964 & 10.964 & 10.964 & 10.964
             & 68.608 & 68.608 & 68.608 & 68.608 & 68.608 \\
    \bottomrule
  \end{tabular}
  }
  \vspace{2mm}
\end{table}

\clearpage
\subsection{Analysis of Evaluation Time} \label{app:benchtime}

In this section, we present a detailed analysis of evaluation time costs for each RDB-to-graph modeling method, which complements the results summarized in Figure~\ref{fig:benchmark_time} in the main paper.
Details are provided in Table~\ref{tab:timeanalysis}.

\begin{table}[ht]
  \centering
  \vspace{-3mm}
  \caption{ Analysis of evaluation costs for each method. Total time is computed as the sum of evaluation time and run time, encompassing on-the-fly evaluation and search method costs. All times are averaged over 10 runs. }
  \vspace{3mm}
  \label{tab:timeanalysis}
  \setlength{\tabcolsep}{4pt}
  \small
    \begin{tabular}{l|ccc}
    \toprule
    \textbf{Method} & \textbf{Total Time (s)} & \textbf{Run Time (s)} & \textbf{Ratio (Total/Run)} \\
    \midrule
    \midrule
    Random & 319016.09 & 0.59 & 540705.24  \\
    Greedy Forward Search (GF) & 104246.09 & 0.39 & 267297.67 \\
    Greedy Backward Search (GB) & 216032.31 & 0.51 & 423592.76  \\
    Greedy Local Search (GL) & 284439.64 & 0.34 & 836587.18 \\
    Evolutionary Algorithm (EA) & 696925.57 & 132.62 & 5255.06 \\
    Bayesian Optimization (BO) & 595853.52 & 1244.17 & 478.92 \\
    Reinforcement Learning (RL) & 753865.31 & 17.66 & 42687.73 \\
    LLM baseline (LLM)  & 50861.62 & 3588.22 & 14.17 \\
    LLM-CoT baseline (CoT)  & 28847.97 & 2924.82 & 9.86 \\
    \bottomrule
    \end{tabular}
\end{table}

\vspace{-3mm}
\subsection{Analysis of LLM Interpretability}
Our analysis shows that interpretability can be enhanced when the LLM-CoT baseline provides explicit reasoning traces for each selected graph transformation action.
During the action planning phase, we prompted the LLM to explain its rationale behind each operation, allowing us to capture structured insights about how it interprets temporal and structural dependencies.

As shown below, the LLM did not merely select from predefined actions but offered justifications referencing temporal order and causal consistency.
These intuitive rationales demonstrate that the LLM’s reasoning process is consistent with the underlying graph structure, helping us understand why each transformation was made.

\begin{lstlisting}[style=textblock]
<reasoning>
1. Current State Analysis:
- The last action restored the FK-PK relationship between results and races, which improved performance slightly
- This suggests that temporal relationships between tables are important for DNF prediction
- The key tables for DNF prediction appear to be results, qualifying, races, and drivers

2. Examining Remaining Relationships:
- The FK relationship between standings and drivers was removed
- A driver's past performance and championship position could correlate with their risk-taking behavior or car reliability

3. Proposed Action:
- Restore the FK-PK relationship between standings and drivers
- Rationale:
  * Driver standings contain cumulative performance metrics (points, position, wins)
  * The temporal nature of standings data could help capture driver form and reliability trends
  * Similar to how restoring race relationships improved performance, driver standings relationship might capture important patterns

4. Expected Impact:
- Should improve model's ability to capture driver-specific patterns over time
- Aligns with the positive results seen from restoring other temporal relationships

Therefore, I recommend using add_fk_pk_edge to restore the relationship between standings and drivers.
</reasoning>
\end{lstlisting}

\clearpage

\section{Further Discussions}
\label{sec:appendix_furtherdiscussions}
\subsection{Limitations}\label{app:limitations}
\textbf{Graph Design Space Constraints.} As described in Section~\ref{subsec:searchspace} of the main paper, we define the graph design space based on the choices in selecting foreign key relations and modeling rows as nodes or edges.
However, additional potential design choices, such as creating dummy tables~\cite{wang20244dbinfer} and advanced feature engineering techniques, are currently excluded due to computational limitations. 
Expanding the graph design space can present a promising direction for future work.

\textbf{GNN Dependence.}
Currently, \ourdata provides precomputed evaluations only for four GNNs: GraphSAGE (sum aggregation), GraphSAGE (mean aggregation), GIN, and GPS. 
Yet, our performance generalization analysis suggests that our findings and the benchmark's utility extend beyond these specific GNNs.

\textbf{Limited Task Coverage.}
While our original goal is to cover all tasks available in RelBench~\cite{robinson2024relbench}, we have excluded some tasks from \stackex and \trials due to extremely large graph design spaces and high computational costs. 
Future work may extend our dataset by covering more tasks.

\subsection{Broader Impact}\label{app:broaderimpact}
Our analysis using \ourdata confirms the importance of strategic graph modeling for relational databases, highlighting its academic and practical significance.

From a research perspective, by providing a unified and well-structured set of benchmarks, \ourdata significantly reduces the evaluation cost for researchers and facilitates reproducible comparisons across methods.
This enables more efficient and robust validation, thereby accelerating progress in RDB-to-graph modeling research.

Furthermore, leveraging \ourdata can significantly enhance RDB-to-graph modeling methods, enabling industries such as finance, healthcare, and e-commerce to improve efficiency and predictive performance on critical tasks.
For example, the financial sector can enhance fraud detection accuracy, while the healthcare sector may develop more precise models for predicting patient outcomes, both directly benefiting from optimized graph models.

\subsection{Analysis of GNN Depth}
\label{sec:gnn-depth}
Following the default configuration of RelBench~\cite{robinson2024relbench}, we fixed the number of GNN layers to two in all experiments.
This configuration provides a consistent and computationally efficient setup, and our ablation study further confirms that it is also empirically well-justified. 
As shown in Tables~\ref{tab:depth-performance}-\ref{tab:depth-corr}, the 2-layer GNN achieves the best average performance across the tasks while also exhibiting a strong correlation with the 3-layer GNN in terms of the top and bottom ranked graph configurations.

We adopt this configuration as it offers a consistent and efficient experimental setup.
However, deeper GNNs could potentially better exploit longer relational paths, especially in databases with large-radius schemas. 
Exploring such depth variations remains a promising direction for future work.

\begin{table}[ht]
  \centering
  \caption{Average performance with different GNN depths (1–3 layers).}
  \label{tab:depth-performance}
  \scalebox{0.8}{
    \begin{tabular}{llccc}
    \toprule
    \textbf{RDB} & \textbf{Task Name (Metric)} & \textbf{1-layer} & \textbf{2-layer} & \textbf{3-layer} \\
    \midrule
    \midrule
    \fone & \driverDNF \small{(AUC-ROC ↑)} & 71.05 ± 0.98 & \textbf{71.76 ± 1.39} & 70.87 ± 1.10 \\
    \avito & \userClick \small{(AUC-ROC ↑)} & 64.03 ± 0.85 & \textbf{64.96 ± 1.03} & 64.88 ± 1.34 \\
    \event & \userAttendance \small{(MAE ↓)} & \textbf{0.249 ± 0.011} & \textbf{0.249 ± 0.011} & 0.251 ± 0.010 \\
    \bottomrule
    \end{tabular}
  }
\end{table}

\begin{table}[ht]
  \centering
  \caption{Correlation between 2-layer and 3-layer GNN configurations (top and bottom 10\%).}
  \label{tab:depth-corr}
  \setlength{\tabcolsep}{4pt}
  \scalebox{0.8}{
    \begin{tabular}{llcc}
    \toprule
    \textbf{RDB} & \textbf{Task Name} & \textbf{Pearson} & \textbf{Spearman} \\
    \midrule
    \midrule
    \fone & \driverDNF & 0.851 & 0.816 \\
    \avito & \userClick & 0.852 & 0.805 \\
    \event & \userAttendance & 0.963 & 0.883 \\
    \bottomrule
    \end{tabular}
  }
\end{table}

\clearpage

\section{Prompt Design}
\label{sec:appendix_prompt}
In this section, we describe the prompts used for our LLM-based baseline implementation, introduced in Section~\ref{subsec:benchmarksettings}.

\begin{figure*}[!ht]
    \centering
    \includegraphics[width=0.95\linewidth]{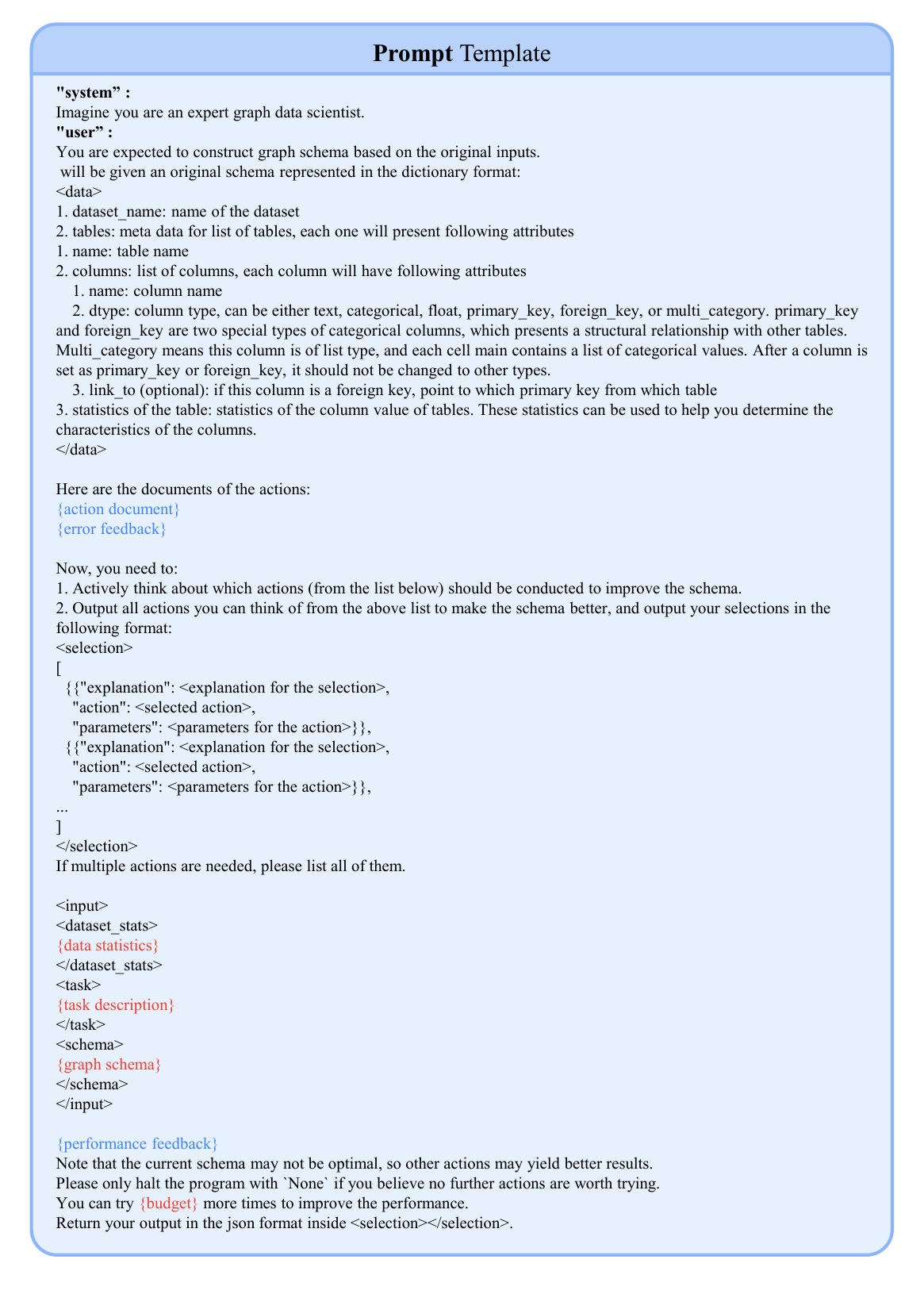}
    \label{fig:templateprompt}
    \caption{Prompt template for main process.}
\end{figure*}

\begin{figure*}[!ht]
    \centering
    \includegraphics[width=0.95\linewidth]{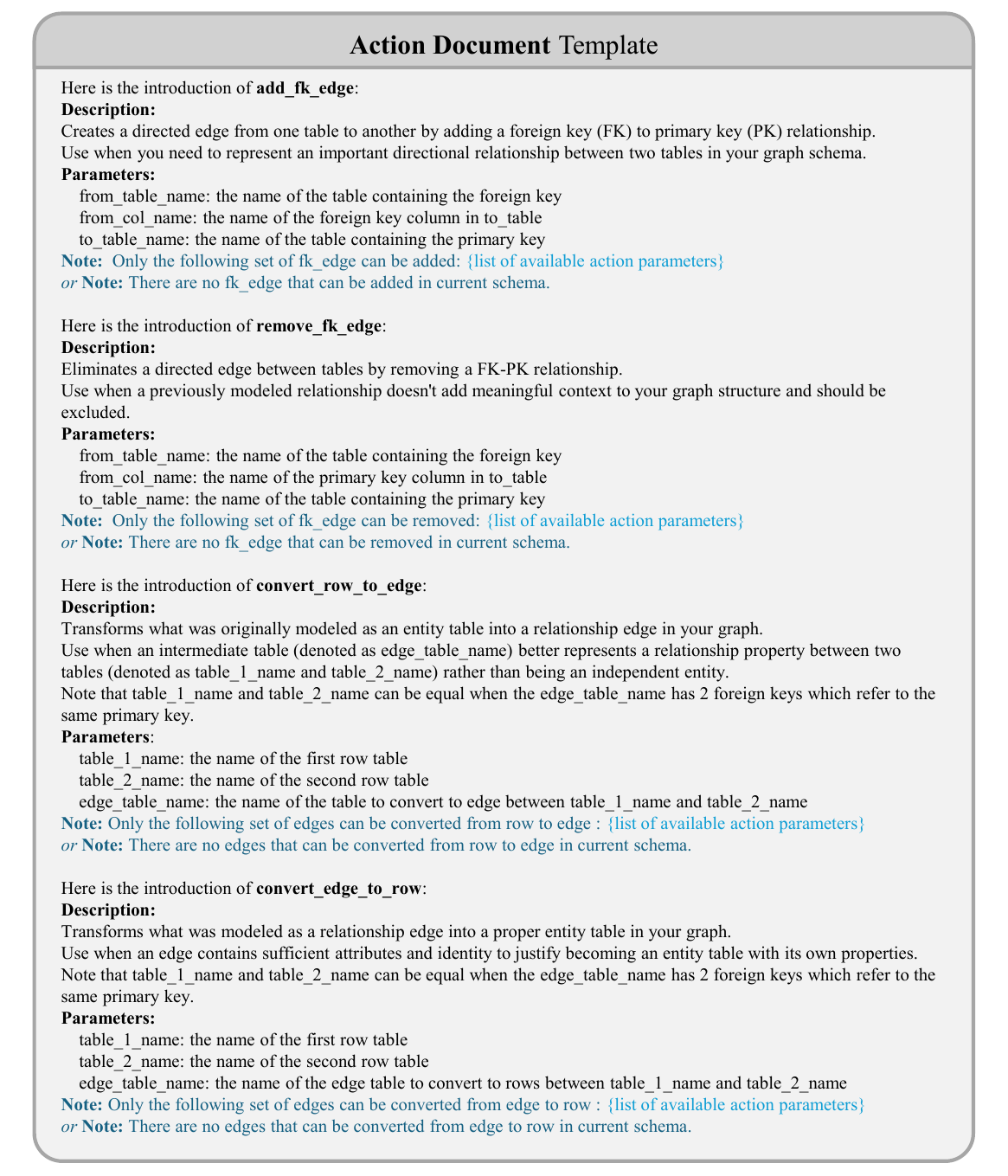}
    \label{fig:templateaction}
    \caption{Prompt template for action document.}
\end{figure*}

\begin{figure*}[!ht]
    \centering
    \includegraphics[width=0.95\linewidth]{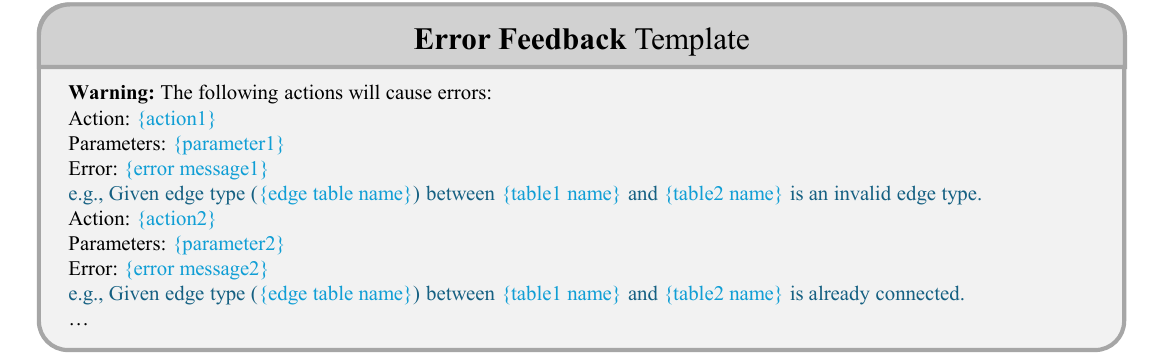}
    \label{fig:templateerrorfeedback}
    \caption{Prompt template for error feedback.}
\end{figure*}

\begin{figure*}[!ht]
    \centering
    \includegraphics[width=0.95\linewidth]{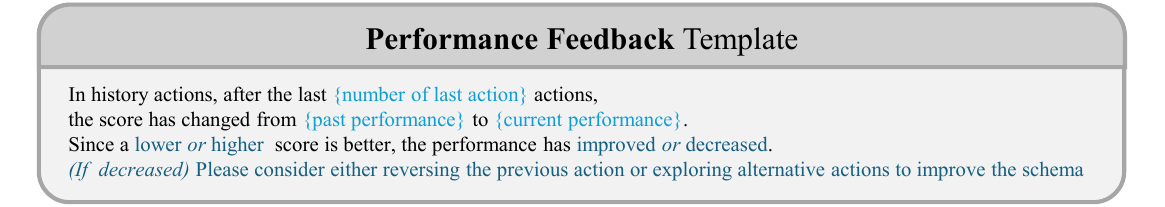}
    \label{fig:templatefeedback}
    \caption{Prompt template for performance feedback.}
\end{figure*}

\begin{figure*}[!ht]
    \centering
    \includegraphics[width=0.95\linewidth]{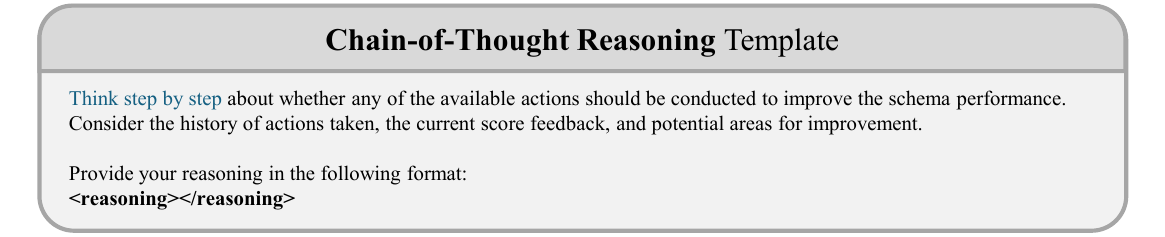}
    \label{fig:templatecot}
    \caption{Prompt template for Chain-of-Thought reasoning.}
\end{figure*}

\end{document}